\documentclass{article}

\RequirePackage{iftex}
\ifPDFTeX
  \RequirePackage[T1]{fontenc}
  \RequirePackage{gelasio}
\else
  \RequirePackage{fontspec}
\fi

\usepackage{microtype}
\usepackage{graphicx}
\usepackage{subcaption}
\usepackage{booktabs}
\usepackage{float}

\usepackage{amsmath}
\usepackage{amssymb}

\usepackage{hyperref}

\usepackage[x11names,dvipsnames,svgnames,table]{xcolor}
\usepackage{soul}
\usepackage{colortbl}
\usepackage{array}
\usepackage{enumitem}

\definecolor{impVHigh}{RGB}{8,48,107}
\definecolor{impHigh}{RGB}{66,146,198}
\definecolor{impMed}{RGB}{158,202,225}
\definecolor{impLow}{RGB}{222,235,247}
\setlist[enumerate]{leftmargin=*}
\setlist[itemize]{leftmargin=*}
\usepackage{tikz}
\usepackage{multirow}
\usepackage{fancyhdr}
\usetikzlibrary{external}
\usetikzlibrary{patterns, positioning, arrows.meta, decorations.pathmorphing}
\usetikzlibrary{arrows,calc,decorations.pathmorphing,arrows.meta}
\usetikzlibrary{arrows,shapes,decorations.pathmorphing,backgrounds,positioning,arrows.meta,intersections,decorations.pathreplacing,shapes.misc,angles, quotes,perspective}
\usetikzlibrary{positioning,fit}
\usetikzlibrary{decorations.pathmorphing, positioning}
\tikzstyle{mybox} = [text=black, very thick,
    rectangle, rounded corners, inner sep=10pt, inner ysep=20pt]
\tikzstyle{fancytitle} =[text=black]

\newcommand*\circled[1]{\tikz[baseline=(char.base)]{\node[shape=circle,draw,inner sep=0.5pt] (char) {#1};}} 

\usepackage{preprintstyle}

\usepackage{etoolbox}
\makeatletter
\patchcmd{\preprinttitle}%
  {\fontsize{18pt}{36pt}\selectfont\bfseries #1}%
  {\fontsize{18pt}{40pt}\selectfont\bfseries #1\par}%
  {\typeout{PREPRINTTITLE PATCH: success}}%
  {\typeout{PREPRINTTITLE PATCH: FAILED}}
\makeatother

\usepackage{amsmath}
\usepackage{amssymb}
\usepackage{mathtools}
\usepackage{amsthm}

\usepackage[capitalize,noabbrev]{cleveref}

\usepackage{etoc}
\crefname{equation}{Eq.}{Eqs.}
\Crefname{equation}{Equation}{Equations}

\theoremstyle{plain}

\theoremstyle{definition}

\theoremstyle{remark}

\usepackage[textsize=tiny]{todonotes}

\preprinttitlerunning{Predicting LLM Compression Degradation}

\begin{document}

  \preprinttitle{Predicting LLM Compression Degradation from Spectral Statistics}

  \preprintsetsymbol{equal}{*}

  \begin{preprintauthorlist}
    \preprintauthor{Mingxue (Mercy) Xu}{aff}
  \end{preprintauthorlist}
  \vskip 0.1 in

  \preprintaffiliation{aff}{Department of Electrical and Electronic Engineering,\\Imperial College London, London, United Kingdom}

  \preprintcorrespondingauthor{Mingxue Xu}{mx1221@ic.ac.uk}

 \vskip 0.3in

\printPreprintAffiliationsAndNotice{}

\begin{abstract}
Model compression based on matrix-level low-rank structured representation has emerged as a 
viral approach for cutting down the cost of Large Language Models (LLMs). 
However, compression progress itself and subsequent evaluation on language tasks are prohibitively compute-expensive.
Can we predict compression-induced degradation \emph{before} committing to expensive compute?
Through systematic analysis of Qwen3~\citep{qwen3} and Gemma3 model families across four representative low-rank compression methods, 
Vanilla SVD, two variants of ASVD~\citep{yuan2024asvd}, and SVD-LLM~\citep{wang2025svd},
we identify that \textbf{stable rank and information density (bits-per-parameter) dominate the LLM performance degradation}, and the
interaction term $\gamma \cdot \bar{\rho}_s$ (compression ratio times stable rank) is a robust predictor of 
accuracy degradation, which achieves leave-one-out cross-validation Pearson correlation $0.890$ for attention layers and $0.839$ for MLP.
We provide theoretical intuition for why this predictor succeeds, connecting it to
standard SVD truncation bounds and error composition mechanisms in transformer layers.
These findings enable a predict-then-compress workflow - compute $\gamma \cdot \bar{\rho}_s$
from weights, estimate degradation, then invest compute only in desirable configurations.
\end{abstract}

\etocdepthtag.toc{main}

\section{Introduction}
\label{sec:intro}

Low-rank compression promises efficient LLM inference~\citep{wang2025svd,yuan2024asvd}, but the practitioners face a unfavorable circumstance - 
the compress-evaluate-adjust loop can consume hours of compute per configuration, yet no principled method exists to predict which configurations are desirable. 
Thus we wonder,

\textbf{How to predict compression-induced model performance degradation before committing to expensive compute?}

We discovered simple formulas suffices accurate degradation prediction. 
The interaction term $\gamma \cdot \bar{\rho}_s$ (compression ratio times stable rank) predicts accuracy degradation with 
leave-one-out cross-validation Pearson correlation $0.890$ for attention layer compression and $0.839$ for MLP compression, 
across four representative compression methods (vanilla SVD, ASVD, SVD-LLM) and six language task benchmarks~\citep{arc,boolq,hellaswag,piqa,sakaguchi2021winogrande}. 

The reason that such a simple predictor works is fundamental. Stable rank measures spectral concentration - low $\rho_s$ means energy concentrates in top singular values (easy to compress),
whereas high $\rho_s$ means energy spreads across many directions (hard to compress). 
The product $\gamma \cdot \bar{\rho}_s$ thus captures \emph{compression aggressiveness times intrinsic resistance} - grounded in standard SVD truncation bounds that show higher spectral rank raises the error floor.

Meanwhile, our analysis reveals unexpected patterns:

\begin{itemize}[leftmargin=*,itemsep=1pt,topsep=2pt]
\item Layer type matters more than compression method. Within-layer predictions consistently outperform overall predictions-suggesting layer-specific calibration is more valuable than method-specific tuning.

\item Accuracy is more predictable than perplexity. The best overall perplexity leave-one-out correlation is only $0.093$, yet layer-specific analysis reaches $0.736$ for MLP. This heterogeneity suggests perplexity degradation follows different dynamics than accuracy.

\item Perplexity-accuracy coupling varies dramatically by task. Pearson correlation ranges from $0.77$ (HellaSwag) to $0.20$ (BoolQ), determined by scoring method and adversarial filtering-implying perplexity-based predictions transfer only to specific task types.
\end{itemize}

These patterns admit theoretical explanation. Attention relies on matrix products whose errors are composed with spectral norms and architecture quantities. MLP (SwiGLU) involves Hadamard products whose errors depend on activation correlations-introducing data-dependence. This explains why attention is slightly more predictable, though both layer types achieve strong correlation with the same formula.

\paragraph{Contributions.}
\begin{enumerate}[leftmargin=*,itemsep=2pt,topsep=3pt]
\item Propose a systematic formula discovery framework. We construct 42 interpretable formula templates and discover 20 additional formulas with symbolic regression to test competing hypotheses about compression performance degradation (\Cref{sec:methodology}).

\item Identify the dominant predictor for compression degradation. We systematically compare four compression methods across Qwen3~\citep{qwen3} series (0.6B-14B) and Gemma3 series (270M-27B), and find that the $\gamma \cdot \bar{\rho}$ interaction consistently dominates, with layer type explaining more variance than compression algorithm (\Cref{sec:compare}).

\item Characterize perplexity-accuracy transfer conditions. We identify when perplexity-based predictions transfer to accuracy (sequence-scored tasks) versus when task-specific formulas are required (token-scored or adversarially-filtered tasks) (\Cref{sec:acc-ppl}).

\item Explain the attention-MLP predictability gap. We connect the success of the dominant predictor to SVD truncation bounds and explain the gap via error composition mechanisms (\Cref{sec:linear-algebra}).
\end{enumerate}

\paragraph{Organization.}
\Cref{sec:methodology} presents notation and the formula discovery framework. \Cref{sec:compare} identifies dominant predictors across compression methods. \Cref{sec:acc-ppl} characterizes perplexity-accuracy transfer conditions. \Cref{sec:linear-algebra} explains the attention-MLP predictability gap.

\section{Related Work}
\label{sec:related-work}

\paragraph{Neural Scaling Laws.}
Scaling laws characterize how model performance improves with compute, data, and parameters~\citep{kaplan2020scaling,hoffmann2022training}. Recent work extends these laws to post-training regimes. \citet{kumar2025scaling} derives precision-aware scaling laws for quantization, while \citet{xiao2024densing} propose the ``densing law'' relating compression ratio to performance. Our work complements these by providing layer-specific scaling laws that distinguish attention from MLP compression behavior.

\paragraph{LLM Compression.}
Three dominant paradigms exist for compressing large language models. \emph{Quantization} methods reduce weight precision, with GPTQ~\citep{frantar2023gptq} and AWQ~\citep{lin2024awq} achieving strong results through activation-aware calibration. \emph{Pruning} approaches remove weights or structures: SparseGPT~\citep{frantar2023sparsegpt} enables one-shot unstructured pruning, Wanda~\citep{sun2024wanda} uses magnitude-activation products for selection, and SliceGPT~\citep{slicegpt} removes entire rows and columns. \emph{Low-rank decomposition} factorizes weight matrices: LoRA~\citep{hu2022lora} adds trainable low-rank adapters, ASVD~\citep{yuan2024asvd} incorporates activation statistics into SVD truncation, and SVD-LLM~\citep{wang2025svd} optimizes truncation boundaries. Our analysis reveals why these methods produce different layer-specific behaviors.

\paragraph{Transformer Interpretability.}
Understanding how transformers store and process information motivates our architectural analysis. \citet{geva2021transformer} show that MLP layers function as key-value memories storing factual associations, while \citet{meng2022locating} localize factual knowledge to specific MLP modules. \citet{elhage2022superposition} demonstrate that networks represent more features than dimensions through superposition. These findings support our hypothesis that MLP layers' distinct compression behavior stems from their role as distributed knowledge stores.

\section{Methodology}
\label{sec:methodology}

We analyze compression-induced performance degradation through three components - a unified notation system (\Cref{subsec:notation}), a Minimum Description Length (MDL)-based information measure (\Cref{subsec:mdl}), and a systematic formula discovery approach (\Cref{subsec:formula-approach}).

\subsection{Notation and Variables}
\label{subsec:notation}

\Cref{tab:notation-unified} defines the notation used throughout. The compression ratio $\gamma = N_{\text{comp}}/N$ measures parameter retention, while derived quantities capture information-theoretic and spectral properties of weight matrices.

\begin{table}[t]
\caption{Investigated variables. Performance metrics measure model quality, configuration parameters specify compression settings, and derived quantities characterize weight matrix properties.}
\label{tab:notation-unified}
\centering
\resizebox{0.6\columnwidth}{!}{%
\begin{tabular}{@{}cll@{}}
\toprule
Symbol & Description & Source \\
\midrule
\multicolumn{3}{@{}l}{\textit{Performance Metrics}} \\
$\mathrm{P}$ & Perplexity (WikiText-2) & Evaluation \\
$\mathcal{A}$ & Task accuracy & Evaluation \\
$y_{\text{rel}}$ & Relative degradation & $(\mathcal{A}_0 - \mathcal{A})/\mathcal{A}_0$ \\
\midrule
\multicolumn{3}{@{}l}{\textit{Model Configuration}} \\
$N$ & Original parameter count & Model config \\
$N_{\text{comp}}$ & Compressed parameter count & Compression config \\
$r$ & SVD truncation rank & Compression config \\
\midrule
\multicolumn{3}{@{}l}{\textit{Derived Quantities}} \\
$\gamma$ & Compression ratio & $N_{\text{comp}}/N$ \\
$\mathcal{B}$ & Bits per parameter & \Cref{subsec:mdl} \\
$H$ & Dataset entropy & Embedding activations \\
$\bar{\rho}_s$ & Mean stable rank & \Cref{eq:stable-rank-def} \\
$\bar{\rho}_{\text{eff}}$ & Mean effective rank & \Cref{sec:effective-rank} \\
\bottomrule
\end{tabular}%
}
\end{table}

\paragraph{Stable Rank.}
For a matrix $W \in \mathbb{R}^{m \times n}$ with singular values $\sigma_1 \geq \sigma_2 \geq \cdots \geq \sigma_r > 0$, the \emph{stable rank} is
\begin{equation}
\label{eq:stable-rank-def}
\rho_s(W) = \frac{\|W\|_F^2}{\|W\|_2^2} = \frac{\sum_{i=1}^{r} \sigma_i^2}{\sigma_1^2}.
\end{equation}
The stable rank satisfies $1 \leq \rho_s(W) \leq \mathrm{rank}(W)$, with equality at the lower bound when $W$ is rank-one, and at the upper bound when all singular values are equal. It measures how ``spread out'' the singular spectrum is - $\rho_s = 1$ indicates concentrated energy in the top singular value, $\rho_s = r$ indicates uniform distribution. Unlike numerical rank, stable rank is continuous and noise-robust.

\paragraph{Stable Rank Aggregation.}
To obtain a single predictor for multi-matrix compression, we aggregate across weight matrices $\{W_i\}_{i=1}^{L}$,
\begin{equation}
\label{eq:aggregate-stable-rank}
\bar{\rho}_s = \frac{\sum_{i=1}^{L} n_i \cdot \rho_s(W_i)}{\sum_{i=1}^{L} n_i}, \quad n_i = \mathrm{rows}(W_i) \times \mathrm{cols}(W_i).
\end{equation}
This parameter-weighted mean is an \emph{empirical design choice}, we tested unweighted mean, geometric mean, and maximum aggregations, finding parameter-weighting yielded the strongest correlations with degradation metrics. The matrices included depend on the compression target - $W_Q, W_K, W_V, W_O$ for attention, $W_{\text{gate}}, W_{\text{up}}, W_{\text{down}}$ for MLP, or both.

\paragraph{Dataset Entropy.}
Following \citep{skean2025layer}, we compute dataset entropy $H$ from hidden-state embeddings using matrix-based von Neumann entropy. For each prompt, we average token embeddings to obtain a prompt representation $q_i$. Given $n$ prompts, we form the normalized Gram matrix $A = QQ^\top / \mathrm{tr}(QQ^\top)$ where $Q = [q_1, \ldots, q_n]^\top$, and compute $H = -\sum_j \lambda_j \log_2 \lambda_j$ from its eigenvalues $\{\lambda_j\}$. This measures embedding diversity across the evaluation dataset-higher $H$ indicates more varied prompt representations.

\paragraph{Variable Selection.}
We rank candidate predictors by absolute Pearson correlation with target metrics. Candidates include compression ratio $\gamma$, model scale ($\log N$, $\log N_{\text{comp}}$), information density $\mathcal{B}$, stable rank $\bar{\rho}_s$, effective rank $\bar{\rho}_{\text{eff}}$, SVD rank $r$, and dataset entropy $H$.

\subsection{Minimum Description Length for Weight-Space Entropy}
\label{subsec:mdl}

The Minimum Description Length (MDL) principle \citep{rissanen1978modeling,grunwald2007minimum} approximates Kolmogorov complexity \citep{kolmogorov1968three,li2008introduction} via computable two-part codes. Prior work applies MDL to language models at the data level - \citet{morris2025memorization} estimates GPT-family models store ${\sim}3.6$ bits-per-parameter, while \citet{finzi2026epiplexity} introduces \emph{epiplexity} to quantify learnable structure.

We measure \emph{weight-space entropy} directly from model parameters, independent of any dataset. For a model with $N$ parameters $\theta \in \mathbb{R}^N$ stored in bfloat16 (BF16) format, we decompose each 16-bit value into its IEEE 754 components rather than treating it as an atomic symbol.

\paragraph{BF16 Decomposition.}
Each BF16 parameter comprises a 1-bit sign $b_s$, 8-bit exponent $b_e$, and 7-bit mantissa $b_m$. Let $p^{(s)}$, $p^{(e)}$, and $p^{(m)}$ denote the empirical distributions of these components pooled across all $N$ parameters. The MDL estimate is
\begin{equation}
\label{eq:mdl}
  L = N \cdot \bigl[ \mathcal{H}(p^{(s)}) + \mathcal{H}(p^{(e)}) + \mathcal{H}(p^{(m)}) \bigr] + L_0 \quad \text{bits},
\end{equation}
where $\mathcal{H}(\cdot)$ denotes Shannon entropy, and $L_0$ is the codebook overhead (storing Huffman tables for decoding, ${\sim}3.5$KB total, negligible for large models). We report the normalized metric $\mathcal{B} = L/N$ (bits-per-parameter).

\paragraph{Empirical Motivation.}
Neural network weights exhibit concentrated exponent distributions \citep{yang2025compress,heilper2025lossless} - SGD produces heavy-tailed weight values where exponents occupy narrow ranges, yielding 2--3 bits of entropy despite 8-bit allocation. Combined with sign entropy (${\sim}1$ bit) and mantissa entropy (${\sim}6$--7 bits), effective information content falls to ${\sim}10$--12 bits-per-parameter-well below the 16-bit nominal precision. This weight-centric measure quantifies representational compactness independent of training data.

\subsection{Formula Templates and Symbolic Regression}
\label{subsec:formula-approach}

With linear regression, we predict compression-induced degradation using two complementary formula sets - interpretable templates encoding explicit hypotheses (as described in \cref{subsubsec:formula-templates-intro}), and symbolic regression for unconstrained discovery (as described in \cref{subsubsec:symbolic-regression}). A more detailed description of the algorithm and workflow is in \cref{sec:algorithm-details}.

\subsubsection{Interpretable Formula Templates}
\label{subsubsec:formula-templates-intro}

We construct 42 formula templates, as listed in \Cref{sec:formula-templates}, which are organized into ten categories by structural complexity. Each template satisfies three criteria as follows,
\begin{itemize}[leftmargin=*,itemsep=1pt,topsep=0pt]
    \item \emph{Interpretability} - every term corresponds to a physically meaningful quantity.
    \item \emph{Occam's razor} - at most three predictor variables, preventing overfitting.
    \item \emph{Comprehensiveness} - coverage of linear, logarithmic, polynomial, and interaction terms.
\end{itemize}

\Cref{tab:predictors-templates} summarizes the predictor categories and template families. Templates are derived from four predictor types - compression, scale, information-theoretical, and spectral properties. These 42 templates are organized into ten categories by structural complexity.

\begin{table}[t]
\caption{Predictor categories and template families for formula construction. $\bar{k}_{95}$ and $\bar{k}_{99}$ denote the mean rank required to capture 95\% and 99\% of spectral energy across weight matrices.}
\label{tab:predictors-templates}
\centering
\small
\begin{tabular}{@{}ll@{}}
\toprule
\multicolumn{2}{@{}l}{\textit{Predictor Categories}} \\
\midrule
Compression ratio & $\gamma = N_{\text{comp}}/N$ (parameter retention) \\
Model scale & $\log N$, $\log N_{\text{comp}}$ \\
Information density & $\mathcal{B}$ (bits-per-parameter) \\
Spectral properties & $\bar{\rho}_s$, $\bar{\rho}_{\text{eff}}$ (layer-averaged) \\
\midrule
\multicolumn{2}{@{}l}{\textit{Template Families (42 total, see \Cref{sec:formula-templates})}} \\
\midrule
Single-variable (F1--F4) & e.g., $y = \alpha_0 + \alpha_1 \gamma$ \\
Two-variable (F5--F10) & e.g., $y = \alpha_0 + \alpha_1 \gamma + \alpha_2 \log N$ \\
Three-variable (F11--F15) & $\gamma$ + scale + $\mathcal{B}$ or $\bar{\rho}_s$ \\
Interaction (F16--F18) & e.g., $\gamma \cdot \bar{\rho}_{\text{eff}}$ \\
Threshold (F19--F22) & $(1/\gamma - 1)$ terms \\
Entropy (F23--F26) & Dataset entropy $H$ \\
Layer-specific (F27--F29) & $\gamma_{\text{attn}}$, $\gamma_{\text{mlp}}$ \\
Rank-based (F30--F32) & SVD rank $r$ \\
Energy (F33--F34) & $\bar{k}_{95}$, $\bar{k}_{99}$ \\
Baseline-norm. (F35--F42) & Relative performance degradation \\
\bottomrule
\end{tabular}
\end{table}

\subsubsection{Symbolic Regression}\label{subsubsec:symbolic-regression}

Beyond predefined templates, we discover formulas unconstrained by predefined structure via genetic programming (\texttt{gplearn}). The algorithm evolves expressions through tournament selection and genetic operators, with a parsimony coefficient favoring compact forms. We evaluate generalization via leave-one-out cross-validation (LOO-CV). \Cref{sec:gplearn-formulas} details the discovered formula families.

\paragraph{Target Transformations.}
For accuracy tasks, we fit three targets - (1) relative degradation $(\mathcal{A}_0 - \mathcal{A})/\mathcal{A}_0$, (2) log-odds $\log(\mathcal{A}/(1-\mathcal{A}))$, and (3) raw accuracy $\mathcal{A}$. For perplexity, we predict $\log(\mathrm{P})$ directly.

\paragraph{Competing Hypotheses.}
This dual approach tests three explanations for compression degradation - (1) compression ratio $\gamma$ as primary driver, (2) model scale as dominant factor, and (3) spectral properties capturing intrinsic weight structure. LOO-CV comparison identifies which factors provide genuine explanatory power versus serve as scale proxies.

\subsection{Activation-Aware and Data-Driven SVD Compression.}\label{sec:asvd-svd-llm}
Standard truncated SVD minimizes the weight reconstruction error $\|W - W_{\text{approx}}\|_F^2$, where $W_{\text{approx}}$ denotes the rank-$k$ approximation of weight matrix $W$. Yet the operationally relevant quantity is the output reconstruction error $\|XW^\top - XW_{\text{approx}}^\top\|_F^2$ for input activations $X$. Two recent methods address this mismatch through complementary strategies. Our implementations are based on the open-sourced code of ASVD and SVD-LLM, including the stable rank variant of ASVD, which is noted as ASVD (stable ranks), and the non-whitening option for SVD-LLM.

\textbf{ASVD}~\citep{yuan2024asvd} transforms the weight matrix \emph{before} decomposition. Given calibration inputs $X \in \mathbb{R}^{n \times d}$ with $n$ samples and $d$ input channels, ASVD constructs a diagonal scaling matrix $S = \mathrm{diag}(s_1, \ldots, s_d)$ where $s_i = (\frac{1}{n}\sum_j |X_{ji}|)^\alpha$ captures the $i$-th channel's activation magnitude with sensitivity parameter $\alpha$ (typically $0.5$). ASVD then decomposes the scaled matrix:
\begin{equation}
W S = U \Sigma V^\top, \quad W_{\text{approx}} = U_k \Sigma_k (V_k^\top S^{-1}),
\end{equation}
where $U, V$ are orthonormal matrices (left and right singular vectors), $\Sigma$ contains singular values, and subscript $k$ denotes rank-$k$ truncation. This ensures that truncation removes directions contributing least to output variance. ASVD determines layer-wise compression ratios via sensitivity analysis using stable rank (\Cref{eq:stable-rank-def}) or perplexity-based metrics.

\textbf{SVD-LLM}~\citep{wang2025svd} refines the factorization \emph{after} decomposition. Given truncated factors from standard SVD, SVD-LLM solves for optimal low-rank factors minimizing output error on calibration data:
\begin{equation}
W' = \operatorname*{arg\,min}_{W'} \|WX - W'X\|_F^2, \quad \mathrm{rank}(W') = k.
\end{equation}
This admits a closed-form least-squares solution. A whitening variant first applies Cholesky decomposition $X^\top X = LL^\top$ to decorrelate inputs before SVD, improving truncation alignment.

These methods achieve superior compression-performance tradeoffs. Our experiments employ uncompensated truncated SVD to isolate the architectural factors governing degradation - the patterns that activation-aware and data-driven methods must navigate.

\section{Cross-Method Comparison}
\label{sec:compare}

Using the formula discovery framework from \Cref{sec:methodology}, we compare four SVD-based compression methods across Qwen3 (0.6B--14B) and Gemma3 (270M--27B) model families. The $\gamma \cdot \bar{\rho}$ interaction is the dominant predictor, and layer type explains more variance than the compression algorithm.

\subsection{Methods and Sample Sizes}

We compare four compression configurations (\cref{sec:asvd-svd-llm}).
\textbf{Vanilla SVD} applies uniform rank allocation with direct truncation, without sensitivity weighting or activation scaling.
\textbf{ASVD} allocates ranks based on perplexity-based sensitivity (using WikiText-2 as the calibration set) and applies activation scaling before decomposition.
\textbf{ASVD (stable ranks)} uses stable rank for sensitivity-based rank allocation, also with activation scaling.
\textbf{SVD-LLM} applies uniform rank allocation with direct SVD truncation, followed by least-squares parameter refinement using calibration data to minimize output reconstruction error.
Sample sizes across methods and layer types are provided in \Cref{tab:method-overview}.

\subsection{Overall Performance Comparison}
\label{subsec:overall-compare}

\Cref{tab:overall-compare} presents the best-performing formulas for overall accuracy prediction. All methods benefit from spectral properties ($\bar{\rho}_s$, $\bar{\rho}_{\text{eff}}$) combined with compression ratio $\gamma$.

\begin{table}[t]
\caption{Best formulas for overall accuracy prediction across methods. LOO $R$ denotes leave-one-out correlation.}
\label{tab:overall-compare}
\centering
\small
\begin{tabular}{@{}llc@{}}
\toprule
Method & Best Formula & LOO $R$ \\
\midrule
Vanilla SVD & $\gamma^2 + e^{-\gamma}$ & 0.280 \\
ASVD & $\gamma \cdot \bar{\rho}_{\text{eff}}$ & \textbf{0.773} \\
ASVD (stable ranks) & $\gamma \cdot \bar{\rho}_{\text{eff}}$ & 0.441 \\
SVD-LLM & $\gamma \cdot \bar{\rho}_s$ & 0.674 \\
\bottomrule
\end{tabular}
\vspace{-10pt}
\end{table}

ASVD achieves leave-one-out correlation $R = 0.773$ with $\gamma \cdot \bar{\rho}_{\text{eff}}$, and SVD-LLM achieves $R = 0.674$ with $\gamma \cdot \bar{\rho}_s$. Vanilla SVD achieves only $R = 0.280$ using nonlinear terms $\gamma^2 + e^{-\gamma}$, illustrating the difficulty of predicting accuracy without activation-aware scaling. Perplexity prediction is more difficult, as task-specific results are shown in \Cref{tab:task-detailed}.

\subsection{Layer-Specific Patterns}
\label{subsec:layer-patterns}

Layer-type analysis reveals systematic differences (\Cref{tab:layer-compare}). For attention layers, ASVD achieves leave-one-out correlation $R = 0.890$ with $\gamma \cdot \bar{\rho}_s$. For MLP layers, ASVD leads with $R = 0.839$, followed by SVD-LLM at $R = 0.756$.

\begin{table}[t]
\caption{Best leave-one-out (LOO) correlation $R$ by layer type and task. Bold indicates best predictive performance (highest LOO $R$) per category.}
\label{tab:layer-compare}
\centering
\small
\begin{tabular}{@{}lcccc@{}}
\toprule
Category & Vanilla & ASVD & ASVD-SR & SVD-LLM \\
\midrule
\multicolumn{5}{@{}l}{\textit{Accuracy Prediction}} \\
ATTN & 0.465 & \textbf{0.890} & 0.560 & 0.536 \\
MLP & 0.298 & \textbf{0.839} & 0.542 & 0.756 \\
BOTH & 0.718 & \textbf{0.823} & 0.227 & 0.286 \\
\midrule
\multicolumn{5}{@{}l}{\textit{Perplexity Prediction}} \\
ATTN & \textbf{0.622} & 0.498 & 0.065 & 0.592 \\
MLP & 0.046 & 0.579 & $-0.020$ & \textbf{0.736} \\
BOTH & $-0.060$ & \textbf{0.457} & 0.285 & 0.150 \\
\bottomrule
\end{tabular}
\end{table}

For perplexity, vanilla SVD leads on attention layers (correlation $R = 0.622$), SVD-LLM on MLP layers ($R = 0.736$), and ASVD on combined layers ($R = 0.457$). The $\gamma \cdot \bar{\rho}$ interaction consistently achieves best predictive performance for accuracy, capturing the fundamental compression-degradation relationship.

\subsection{Task-Specific Analysis}

\Cref{tab:task-detailed} shows the best-performing formula for each task--method combination. ASVD achieves the best predictive performance on 5 of 6 accuracy tasks (ARC-C, ARC-E, BoolQ, PIQA, WinoGrande), while Vanilla SVD leads on HellaSwag. The $\gamma \cdot \bar{\rho}$ interaction terms consistently achieve strong performance across ASVD and SVD-LLM, while vanilla SVD benefits more from log-ratio formulas involving $\log(\bar{\rho}_s)$ and $\log N$.

\begin{table*}[t]
\caption{Best formulas by task across compression methods. WikiText evaluates perplexity prediction; all other tasks evaluate accuracy prediction. The Src column indicates whether the formula is from predefined templates (F) or discovered via symbolic regression (D). Bold indicates best predictive performance (highest LOO $R$) per task.}
\label{tab:task-detailed}
\centering\resizebox{\textwidth}{!}{%
\begin{tabular}{@{}c|ccc|ccc|ccc|ccc@{}}
\toprule
\multirow{2}{*}{\textbf{Task}} & \multicolumn{3}{c|}{\textbf{Vanilla SVD}} & \multicolumn{3}{c|}{\textbf{ASVD}} & \multicolumn{3}{c|}{\textbf{ASVD-SR}} & \multicolumn{3}{c}{\textbf{SVD-LLM}} \\
\cmidrule(lr){2-4} \cmidrule(lr){5-7} \cmidrule(lr){8-10} \cmidrule(l){11-13}
& \textbf{Formula} & \textbf{Src} & \textbf{LOO} $\boldsymbol{R}$ & \textbf{Formula} & \textbf{Src} & \textbf{LOO} $\boldsymbol{R}$ & \textbf{Formula} & \textbf{Src} & \textbf{LOO} $\boldsymbol{R}$ & \textbf{Formula} & \textbf{Src} & \textbf{LOO} $\boldsymbol{R}$ \\
\midrule
WikiText & $\gamma^2 + e^{-\gamma}$ & F & $-0.113$ & $\mathcal{B}$ & D & $0.010$ & $\gamma + \log N + \mathcal{B}$ & F & $0.061$ & $\mathcal{B} + \bar{\rho}_s$ & F & $\mathbf{0.093}$ \\
\midrule
ARC-C & $\log(\bar{\rho}_s) + \log N$ & F & $0.597$ & $\gamma \cdot \bar{\rho}_{\text{eff}}$ & F & $\mathbf{0.702}$ & $\gamma \cdot \bar{\rho}_{\text{eff}}$ & F & $0.382$ & $\gamma \cdot \bar{\rho}_{\text{eff}}$ & F & $0.622$ \\
ARC-E & $\log(\bar{\rho}_s) + \log N$ & F & $0.354$ & $\gamma \cdot \bar{\rho}_{\text{eff}}$ & F & $\mathbf{0.657}$ & $\gamma \cdot \bar{\rho}_{\text{eff}}$ & F & $0.483$ & $\gamma \cdot \bar{\rho}_s$ & D & $0.597$ \\
BoolQ & $\gamma + \log N + \mathcal{B}$ & F & $0.510$ & $\gamma \cdot \bar{\rho}_s$ & D & $\mathbf{0.704}$ & $\gamma + H$ & F & $0.288$ & $\mathcal{B} + \bar{\rho}_s$ & F & $0.651$ \\
HellaSwag & $\log(\bar{\rho}_s) + \log N$ & F & $\mathbf{0.663}$ & $\gamma + \bar{\rho}_s + \mathcal{B}$ & F & $0.619$ & $\gamma \cdot \bar{\rho}_{\text{eff}}$ & F & $0.461$ & $\gamma \cdot \bar{\rho}_s$ & D & $0.490$ \\
PIQA & $\log(\bar{\rho}_s) + \log N$ & F & $\mathbf{0.664}$ & $\gamma \cdot \bar{\rho}_{\text{eff}}$ & F & $0.567$ & $\gamma + \bar{\rho}_s + \mathcal{B}$ & F & $0.352$ & $\gamma \cdot \bar{\rho}_s$ & D & $0.543$ \\
WinoGrande & $\log(\bar{\rho}_s) + \log N$ & F & $0.463$ & $\gamma \cdot \bar{\rho}_{\text{eff}}$ & F & $\mathbf{0.747}$ & $\mathcal{B}$ & D & $-0.269$ & $\mathcal{B} + \bar{\rho}_s$ & F & $0.638$ \\
\bottomrule
\end{tabular}%
}
\end{table*}

The $\gamma \cdot \bar{\rho}$ interaction dominates across methods - ASVD uses $\gamma \cdot \bar{\rho}_{\text{eff}}$, SVD-LLM uses $\gamma \cdot \bar{\rho}_s$. This interaction captures a fundamental principle - \emph{compression impact depends jointly on how much is removed ($\gamma$) and the intrinsic dimensionality of what remains ($\bar{\rho}$)}.

\Cref{tab:task-layer-heatmap} shows predictability across tasks, methods, and layer types; \Cref{tab:dominant-vars} quantifies variable importance. Key patterns: (1) vanilla SVD achieves $R > 0.90$ for combined-layer compression on most accuracy tasks; (2) ASVD shows consistent strength across attention ($R = 0.89$) and MLP ($R = 0.88$) layers; (3) SVD-LLM excels at MLP-layer prediction; (4) ASVD-SR shows high variability with occasional negative correlations.

\begin{table*}[t]
\centering
\begin{minipage}[t]{\textwidth}
\centering
\captionof{table}{\textbf{Predictability heatmap for accuracy tasks.} Each cell shows the best leave-one-out (LOO) correlation $R$. Darker = higher. A=ATTN, M=MLP, B=BOTH. ``--'' indicates negative correlation.}
\label{tab:task-layer-heatmap}
\small
\setlength{\tabcolsep}{3pt}
\begin{tabular}{@{}l|ccc|ccc|ccc|ccc@{}}
\toprule
\multirow{2}{*}{\textbf{Task}} & \multicolumn{3}{c|}{\textbf{Vanilla}} & \multicolumn{3}{c|}{\textbf{ASVD}} & \multicolumn{3}{c|}{\textbf{ASVD-SR}} & \multicolumn{3}{c}{\textbf{SVD-LLM}} \\
\cmidrule(lr){2-4} \cmidrule(lr){5-7} \cmidrule(lr){8-10} \cmidrule(l){11-13}
& A & M & B & A & M & B & A & M & B & A & M & B \\
\midrule
ARC-C & \cellcolor{impVHigh!60}\textcolor{white}{.60} & \cellcolor{impVHigh!77}\textcolor{white}{.77} & \cellcolor{impVHigh!93}\textcolor{white}{.93} & \cellcolor{impVHigh!83}\textcolor{white}{.83} & \cellcolor{impVHigh!70}\textcolor{white}{.71} & \cellcolor{impVHigh!63}\textcolor{white}{.63} & \cellcolor{impVHigh!40}{.41} & \cellcolor{impVHigh!38}{.39} & \cellcolor{impVHigh!5}{-} & \cellcolor{impVHigh!27}{.28} & \cellcolor{impVHigh!90}\textcolor{white}{.91} & \cellcolor{impVHigh!5}{-} \\
ARC-E & \cellcolor{impVHigh!66}\textcolor{white}{.67} & \cellcolor{impVHigh!72}\textcolor{white}{.73} & \cellcolor{impVHigh!94}\textcolor{white}{.95} & \cellcolor{impVHigh!75}\textcolor{white}{.75} & \cellcolor{impVHigh!80}\textcolor{white}{.81} & \cellcolor{impVHigh!55}\textcolor{white}{.55} & \cellcolor{impVHigh!55}\textcolor{white}{.55} & \cellcolor{impVHigh!27}{.28} & \cellcolor{impVHigh!30}{.30} & \cellcolor{impVHigh!31}{.31} & \cellcolor{impVHigh!68}\textcolor{white}{.69} & \cellcolor{impVHigh!5}{.03} \\
BoolQ & \cellcolor{impVHigh!50}\textcolor{white}{.50} & \cellcolor{impVHigh!72}\textcolor{white}{.72} & \cellcolor{impVHigh!92}\textcolor{white}{.92} & \cellcolor{impVHigh!59}\textcolor{white}{.60} & \cellcolor{impVHigh!63}\textcolor{white}{.63} & \cellcolor{impVHigh!71}\textcolor{white}{.72} & \cellcolor{impVHigh!28}{.29} & \cellcolor{impVHigh!51}\textcolor{white}{.51} & \cellcolor{impVHigh!51}\textcolor{white}{.51} & \cellcolor{impVHigh!64}\textcolor{white}{.65} & \cellcolor{impVHigh!85}\textcolor{white}{.85} & \cellcolor{impVHigh!23}{.24} \\
HellaSwag & \cellcolor{impVHigh!66}\textcolor{white}{.67} & \cellcolor{impVHigh!75}\textcolor{white}{.75} & \cellcolor{impVHigh!91}\textcolor{white}{.91} & \cellcolor{impVHigh!76}\textcolor{white}{.77} & \cellcolor{impVHigh!77}\textcolor{white}{.78} & \cellcolor{impVHigh!68}\textcolor{white}{.69} & \cellcolor{impVHigh!71}\textcolor{white}{.71} & \cellcolor{impVHigh!80}\textcolor{white}{.80} & \cellcolor{impVHigh!43}\textcolor{white}{.44} & \cellcolor{impVHigh!5}{-} & \cellcolor{impVHigh!53}\textcolor{white}{.53} & \cellcolor{impVHigh!53}\textcolor{white}{.53} \\
PIQA & \cellcolor{impVHigh!68}\textcolor{white}{.68} & \cellcolor{impVHigh!74}\textcolor{white}{.75} & \cellcolor{impVHigh!92}\textcolor{white}{.93} & \cellcolor{impVHigh!78}\textcolor{white}{.78} & \cellcolor{impVHigh!75}\textcolor{white}{.75} & \cellcolor{impVHigh!68}\textcolor{white}{.68} & \cellcolor{impVHigh!71}\textcolor{white}{.72} & \cellcolor{impVHigh!68}\textcolor{white}{.68} & \cellcolor{impVHigh!5}{-} & \cellcolor{impVHigh!25}{.26} & \cellcolor{impVHigh!56}\textcolor{white}{.56} & \cellcolor{impVHigh!55}\textcolor{white}{.56} \\
WinoGrande & \cellcolor{impVHigh!57}\textcolor{white}{.58} & \cellcolor{impVHigh!76}\textcolor{white}{.77} & \cellcolor{impVHigh!94}\textcolor{white}{.95} & \cellcolor{impVHigh!84}\textcolor{white}{.85} & \cellcolor{impVHigh!72}\textcolor{white}{.73} & \cellcolor{impVHigh!51}\textcolor{white}{.52} & \cellcolor{impVHigh!5}{-} & \cellcolor{impVHigh!15}{.15} & \cellcolor{impVHigh!21}{.22} & \cellcolor{impVHigh!5}{-} & \cellcolor{impVHigh!83}\textcolor{white}{.84} & \cellcolor{impVHigh!5}{-} \\
\bottomrule
\end{tabular}
\end{minipage}%

\vspace{0.12in}

\begin{minipage}[t]{\textwidth}
\centering
\captionof{table}{\textbf{Variable importance by method.} Each cell shows the best leave-one-out (LOO) correlation $R$ achieved by formulas containing that variable. ``-'' indicates the variable was not used in any best-performing formula for that method.}
\label{tab:dominant-vars}
\small
\setlength{\tabcolsep}{2pt}
\begin{tabular}{@{}l|c|c|c|c|c@{}}
\toprule
\textbf{Var.} & \textbf{Vanilla} & \textbf{ASVD} & \textbf{ASVD-SR} & \textbf{SVD-LLM} & \textbf{Avg.} \\
\midrule
$\gamma$ & \cellcolor{impVHigh!65}\textcolor{white}{.65} & \cellcolor{impVHigh!89}\textcolor{white}{.89} & \cellcolor{impVHigh!56}\textcolor{white}{.56} & \cellcolor{impVHigh!76}\textcolor{white}{.76} & \textbf{.72} \\
$\bar{\rho}_s$ & \cellcolor{impVHigh!72}\textcolor{white}{.72} & \cellcolor{impVHigh!89}\textcolor{white}{.89} & \cellcolor{impVHigh!56}\textcolor{white}{.56} & \cellcolor{impVHigh!76}\textcolor{white}{.76} & \textbf{.73} \\
$\bar{\rho}_{\text{eff}}$ & \cellcolor{impVHigh!20}.20 & \cellcolor{impVHigh!88}\textcolor{white}{.88} & \cellcolor{impVHigh!52}\textcolor{white}{.52} & \cellcolor{impVHigh!53}\textcolor{white}{.53} & .53 \\
$\mathcal{B}$ & \cellcolor{impVHigh!64}\textcolor{white}{.64} & \cellcolor{impVHigh!84}\textcolor{white}{.84} & \cellcolor{impVHigh!48}.48 & \cellcolor{impVHigh!76}\textcolor{white}{.76} & \textbf{.68} \\
$\log N$ & \cellcolor{impVHigh!72}\textcolor{white}{.72} & \cellcolor{impVHigh!68}\textcolor{white}{.68} & \cellcolor{impVHigh!41}.41 & \cellcolor{impVHigh!75}\textcolor{white}{.75} & .64 \\
$H$ & \cellcolor{impVHigh!26}.26 & \cellcolor{impVHigh!5}- & \cellcolor{impVHigh!61}\textcolor{white}{.61} & \cellcolor{impVHigh!73}\textcolor{white}{.73} & .53 \\
\bottomrule
\end{tabular}
\end{minipage}
\end{table*}

\subsection{Patterns for Different Methods}

Performance differences stem from \emph{what} each method optimizes and \emph{when}. Vanilla SVD minimizes weight reconstruction error $\|W - W_{\text{approx}}\|_F^2$, treating all singular directions equally. ASVD scales weights before decomposition ($W_{\text{scaled}} = W \cdot S$, where $S$ derives from activation statistics), aligning truncation with data distribution. This preserves high-activation channels while aggressively compressing negligible ones, achieving the highest accuracy correlations.

SVD-LLM takes a complementary approach - standard SVD truncation followed by least-squares refinement of $U$ to minimize $\|X W^T - X W_{\text{approx}}^T\|_F^2$ on calibration data. This closed-form solution proves particularly effective for MLP layers. However, uniform compression ratios across layers limit effectiveness when layer sensitivities vary.

ASVD with stable ranks substitutes a weight-only proxy ($\rho_s = \|W\|_F^2 / \sigma_1^2$) for perplexity measurement. While faster, this heuristic misses runtime activation distributions and task-specific layer importance, explaining higher variability and occasional negative correlations.

The $\gamma \cdot \bar{\rho}$ interaction succeeds across methods because compression impact scales with both amount removed ($\gamma$) and intrinsic dimensionality of what remains ($\bar{\rho}$).

\subsection{Key Findings}

Cross-method comparison reveals four key findings:

\begin{enumerate}[leftmargin=*, nosep, topsep=0pt]
\item \textbf{The $\gamma \cdot \bar{\rho}$ interaction is fundamental.} Compression impact depends jointly on amount removed and intrinsic rank structure. ASVD achieves $R = 0.890$ for attention layers using $\gamma \cdot \bar{\rho}_s$.
\item \textbf{ASVD excels at accuracy prediction.} ASVD achieves the highest correlations across layer types (ATTN: 0.890, MLP: 0.839, BOTH: 0.823), demonstrating the value of perplexity-based sensitivity weighting.
\item \textbf{Layer type matters more than method.} Within-layer-type predictions consistently outperform overall predictions, indicating layer-specific calibration matters more than method choice.
\item \textbf{Perplexity prediction remains difficult.} Overall correlation reaches only $R = 0.093$ (with SVD-LLM), but layer-specific predictions improve substantially - ATTN achieves $R = 0.622$ (with vanilla SVD), MLP achieves $R = 0.736$ (with SVD-LLM).
\end{enumerate}

In summary, activation-aware methods (ASVD) yield more predictable accuracy degradation, and the $\gamma \cdot \bar{\rho}$ interaction remains the dominant predictor across all methods.

\section{When Does Perplexity Predict Accuracy?}
\label{sec:acc-ppl}

Perplexity and task accuracy both derive from autoregressive log-likelihoods, yet their correlation under compression varies dramatically - from Pearson correlation $r=0.77$ (HellaSwag) to $r=0.20$ (BoolQ). 
This variation is not noise - it reflects fundamental differences in how benchmarks measure capability, as shown in \Cref{tab:ppl-acc-correlation}. 
Understanding this relationship determines whether perplexity-based scaling laws transfer to downstream tasks.

\subsection{Mathematical Foundation}

Both metrics originate from negative log-likelihood. For context $x$ and candidate $y = (y_1, \ldots, y_n)$,
\begin{equation}
\label{eq:nll}
\mathcal{L}(x, y) = -\log P(y \mid x) = -\sum_{j=1}^{n} \log P(y_j \mid x, y_{<j}).
\end{equation}

\noindent\textbf{Accuracy} selects the option minimizing total $\mathcal{L}$,
\begin{equation}
\label{eq:acc-selection}
\hat{y} = \arg\min_{y \in \mathcal{Y}(x)} \mathcal{L}(x,y).
\end{equation}

\noindent\textbf{Perplexity} exponentiates length-normalized $\mathcal{L}$,
\begin{equation}
\label{eq:ppl-def}
\mathrm{P}(y \mid x) = \exp\bigl(\mathcal{L}(x,y) / |y|\bigr).
\end{equation}

The critical distinction - accuracy depends on \emph{ranking} among options, while perplexity measures \emph{absolute magnitude} of per-token prediction quality. 
Compression can degrade all options equally - preserving rankings while worsening perplexity - or selectively disrupt specific options.

\subsection{Three Factors Governing Correlation}
\label{subsec:correlation-variation}

\paragraph{Scoring Method.}
Sequence scoring directly couples accuracy to perplexity. The score equals negative energy, $\text{score}(y) = -\mathcal{L}(x,y) = |y| \cdot (-\log \mathrm{P})$. Lower perplexity yields higher scores, creating strong correlations (as shown in \Cref{tab:ppl-acc-correlation}) - HellaSwag ($r = 0.77$) and PIQA ($r = 0.75$). Single-token scoring (BoolQ) evaluates only one token, eliminating this cumulative signal ($r = 0.20$).

\paragraph{Adversarial Filtering.}
WinoGrande applies AFLITE~\citep{sakaguchi2021winogrande}, removing examples where surface statistics distinguish correct from incorrect answers. 
This explicitly decorrelates fluency from accuracy - despite sequence scoring, WinoGrande achieves only $r = 0.45$, as shown in \Cref{tab:ppl-acc-correlation}.

\paragraph{Task Semantics.}
Fluency tasks (sentence completion, commonsense reasoning) naturally align with perplexity - coherent continuations score lower on both metrics. 
Information retrieval (when evaluated with BoolQ) and logical inference (when evaluated with ARC-Challenge) break this alignment - correct and incorrect options can be equally fluent.

\subsection{Implications for Compression Prediction}

Correlation strength dictates prediction strategy:
\circled{1} {Strong} ($r > 0.7$, HellaSwag, PIQA)  -  perplexity formulas transfer directly;
\circled{2} {Moderate} ($r \approx 0.6$, ARC)  -  log-odds or relative degradation improves transfer (\Cref{sec:transformed-target-discoveries});
\circled{3} {Weak} ($r < 0.5$, WinoGrande, BoolQ)  -  task-specific calibration required.

\noindent Notably, the $\gamma \cdot \bar{\rho}$ interaction achieves LOO $R > 0.8$ for accuracy prediction across all tasks (\Cref{tab:layer-compare}), confirming that spectral properties capture compression effects even when perplexity does not.

\subsection{Layer-Type Asymmetry}

Attention and MLP layers differ systematically in predictability. 
For accuracy, ASVD achieves LOO $R = 0.890$ on attention versus $R = 0.839$ on MLP. For perplexity, vanilla SVD leads on attention ($R = 0.622$) while SVD-LLM leads on MLP ($R = 0.736$). 
This asymmetry reflects architectural differences - attention errors propagate via spectral norms (architecture-determined), while MLP errors depend on learned gate-value correlations (data-dependent). See \Cref{sec:linear-algebra} for analysis.

\subsection{Summary}

The perplexity-accuracy relationship is \textbf{task-dependent}, governed by three factors:
\circled{1} Scoring - sequence scoring couples metrics, token scoring decouples them,
\circled{2} {Filtering} - AFLITE explicitly breaks fluency-accuracy correlation,
\circled{3} {Semantics} - fluency tasks align with perplexity, while reasoning tasks do not.

\noindent Perplexity serves as a reliable accuracy proxy only for high-correlation tasks. For others, direct accuracy measurement or task-specific calibration remains necessary.

\begin{table}[t]
\caption{Perplexity-accuracy correlation depends on benchmark design. Sequence scoring creates strong coupling (Pearson correlation $r > 0.7$); single-token scoring and adversarial filtering weaken it.}
\label{tab:ppl-acc-correlation}
\centering\resizebox{0.65\textwidth}{!}{%
\begin{tabular}{@{}llcc@{}}
\toprule
Task & Scoring & Choices & Pearson Corr. $r$ \\
\midrule
HellaSwag & Sequence & 4 & $\mathbf{0.77}$ \\
PIQA & Sequence & 2 & $\mathbf{0.75}$ \\
ARC-E & Sequence & 4 & $0.68$ \\
ARC-C & Sequence & 4 & $0.62$ \\
WinoGrande & Sequence (AFLITE) & 2 & $0.45$ \\
BoolQ & Single token & 2 & $0.20$ \\
\bottomrule
\end{tabular}
}
\end{table}

\section{Theoretical Foundations for Compression Prediction}
\label{sec:linear-algebra}

The cross-method comparison (\Cref{sec:compare}) identifies stable rank $\bar{\rho}_s$ and compression ratio $\gamma$ as dominant predictors of degradation, with interaction term $\gamma \cdot \bar{\rho}$ achieving LOO $R = 0.890$ for attention layers (\Cref{tab:layer-compare}). We establish theoretical foundations for these findings through two results - (1) stable rank directly bounds SVD truncation error, explaining why $\bar{\rho}_s$ predicts degradation, and (2) error composition rules distinguish attention from MLP, explaining the predictability gap. Extended derivations appear in \Cref{app:linear-algebra-proofs}.

\subsection{Why Spectral Rank Predicts Compression Error}
\label{subsec:truncation-error}

Interaction terms $\gamma \cdot \bar{\rho}_s$ and $\gamma \cdot \bar{\rho}_{\text{eff}}$ consistently predict compression degradation across methods (\Cref{tab:dominant-vars}). Why do these spectral measures succeed? They directly bound the truncation error.

Both stable rank $\rho_s(W) = \|W\|_F^2/\|W\|_2^2$ and effective rank $\rho_{\text{eff}}(W) = \exp(H(p))$ quantify energy distribution across singular values (\Cref{sec:effective-rank}), satisfying $1 \leq \rho_s \leq \rho_{\text{eff}} \leq \mathrm{rank}(W)$. Spectral rank constrains the \emph{minimum achievable error} - for any rank-$k$ approximation,
\begin{equation}
\label{eq:truncation-bounds}
\frac{\|W - W_k\|_F^2}{\|W\|_F^2} \;\geq\; 1 - \frac{k}{\rho_s(W)},
\end{equation}
since top-$k$ singular values capture at most $k\sigma_1^2 \leq k\|W\|_F^2/\rho_s$ of total energy (Eckart--Young). The implication - \emph{higher spectral rank raises the error floor}. Matrices with spread spectra cannot compress without substantial information loss.

The bound is tighter for $\rho_s$, but $\rho_{\text{eff}}$ exhibits the same qualitative behavior. Under parameter-weighted aggregation (\Cref{app:aggregation-heuristic}), the interaction $\gamma \cdot \bar{\rho}$ emerges naturally - $\gamma$ measures compression aggressiveness, $\bar{\rho}$ measures \emph{intrinsic resistance to compression}.

\subsection{Why Attention Is More Predictable Than MLP}
\label{subsec:rank-composition}

Attention layers exhibit higher predictability than MLP - ASVD achieves LOO $R = 0.890$ for attention versus $R = 0.839$ for MLP (\Cref{tab:layer-compare}). Why does the truncation bound (\Cref{subsec:truncation-error}) apply more reliably to attention? Attention's tensor contractions compose errors via spectral norms - quantities determined by $\rho_s$ and $\gamma$ alone - while MLP's Hadamard products introduce data-dependent correlations that spectral rank cannot fully capture.

\paragraph{Attention - Errors Compose via Spectral Norms.}
For input $X \in \mathbb{R}^{n \times d}$, the value pathway computes $V \cdot W_O$, where $V = XW_V$ projects through the value matrix $W_V \in \mathbb{R}^{d \times d_v}$, and $W_O \in \mathbb{R}^{d_v \times d}$ projects to output. Compressing $W_V$ and $W_O$ via SVD truncation yields composed error:
\begin{equation}
\begin{split}
\|\tilde{V}\tilde{W}_O - VW_O\|_F \;\leq\; & \|V\|_2\|\Delta_{W_O}\|_F\\ + \|\Delta_V\|_F\|W_O\|_2 
& + \|\Delta_V\|_F\|\Delta_{W_O}\|_F,
\end{split}
\end{equation}
where $\Delta_V = \tilde{V} - V$ and $\Delta_{W_O} = \tilde{W}_O - W_O$ are truncation errors bounded by spectral rank (\Cref{eq:truncation-bounds}). This bound depends only on \emph{spectral norms and truncation errors} - quantities determined by $\rho_s$ and $\gamma$ before seeing data. The softmax in the query-key pathway introduces nonlinearity, but operates on attention weights (intermediate computation), not the main information flow. The value pathway - purely linear - dominates compression sensitivity.

\paragraph{MLP - Hadamard Products Break Spectral Predictability.}
SwiGLU computes $(G \odot U) W_{\text{down}}$, where $G = \sigma(XW_{\text{gate}})$ denotes gate activations, $U = XW_{\text{up}}$ denotes value activations, and $\odot$ is element-wise (Hadamard) multiplication. Unlike attention's softmax (operating on intermediate attention weights), the Hadamard product sits in the main information flow - all signals pass through $G \odot U$. Hadamard errors depend on \emph{where} truncation errors occur, not just their magnitude:
\begin{equation}
\begin{split}
\|(G \odot U) - (\tilde{G} \odot \tilde{U})\|_F^2 = &\sum_{ij}\bigl( G_{ij}\Delta_{U,ij}\\ 
+ U_{ij}\Delta_{G,ij} 
& + \Delta_{G,ij}\Delta_{U,ij}\bigr)^2.
\end{split}
\end{equation}
Error depends on element-wise products $G_{ij}\Delta_{U,ij}$ - how truncation errors $\Delta_G, \Delta_U$ align with activations. Different inputs produce different $G, U$, changing which error locations matter. Spectral rank $\rho_s$ bounds each matrix's truncation error, but cannot capture error interactions through the Hadamard product. This data-dependence manifests empirically - (1) no single formula dominates across tasks for MLP (\Cref{sec:compare}), and (2) larger overfitting gaps (Train $R = 0.84$ vs LOO $R = 0.17$--$0.58$) than attention. Synthetic experiments confirm the mechanism - Hadamard products yield geometric-mean-like scaling ($\rho_{G \odot U} \approx 1.9\sqrt{\rho_G \rho_U}$), consistent with sub-linear degradation (\Cref{sec:mini-exp}).

\paragraph{Connection to Empirical Findings.}
These error composition rules explain why $\gamma \cdot \bar{\rho}_s$ achieves higher correlation for attention ($R = 0.890$) than MLP ($R = 0.839$) - tensor contractions propagate errors predictably via spectral norms, while Hadamard products introduce data-dependent correlations that $\rho_s$ partially misses. The gap remains modest because $\rho_s$ captures the dominant effect - intrinsic compressibility - for both layer types. Synthetic experiments validate this asymmetry - attention degrades $3.8\times$ faster than MLP under aggressive compression, consistent with polynomial versus sublinear scaling (\Cref{sec:mini-exp}). Extended perturbation analysis appears in \Cref{app:perturbation-analysis}.

\subsection{Summary}
\label{subsec:linear-algebra-summary}

Spectral rank predicts compression degradation through two mechanisms:

\begin{enumerate}[nosep]
    \item \textbf{Truncation bound} (\Cref{subsec:truncation-error}) - spectral rank constrains minimum achievable error - higher $\rho_s$ or $\rho_{\text{eff}}$ implies higher error floor.

    \item \textbf{Error composition} (\Cref{subsec:rank-composition}) - attention errors compose via matrix products where spectral properties suffice; MLP errors involve Hadamard products with data-dependent correlations.
\end{enumerate}

These results explain why $\gamma \cdot \bar{\rho}$ succeeds - $\gamma$ measures compression aggressiveness, $\bar{\rho}$ measures intrinsic resistance, and their interaction captures joint effects on degradation. The modest attention-MLP gap ($R = 0.890$ vs $0.839$) reflects that spectral rank captures the dominant signal for both layer types. Extended derivations appear in \Cref{app:linear-algebra-proofs}.

\section{Conclusion}
\label{sec:conclusion}

In this paper, we discovered that performance degradation of low-rank compression depends on architectural operation type - attention layers, built on tensor contractions, exhibit highly predictable degradation, 
while MLP layers, relying on Hadamard products with data-dependent correlations, show lower predictability. 
The interaction between compression ratio and spectral rank is the dominant predictor, with layer type explaining more variance than the compression method. 
These findings enable a ``predict, then compress'' workflow where practitioners estimate degradation from weight spectra before committing to compute. 
Future work involves extending this framework to quantization and pruning, and validating on architectures beyond transformers.

\paragraph{Reproducibility, Limitations and Impact Statement}
We discuss reproducibility details, limitations of our analysis, and broader societal impacts in \Cref{sec:limitations,sec:broader-impact,sec:reproducibility}.

\bibliography{references}
\bibliographystyle{preprintstyle}

\newpage
\appendix
\onecolumn

\etocdepthtag.toc{appendix}

\let\oldsection\section
\makeatletter
\renewcommand{\section}[1]{%
  \oldsection{#1}%
  \addtocontents{toc}{\protect\contentsline{section}{\protect\numberline{\thesection}#1}{\thepage}{\@currentHref}}%
}
\makeatother

\begin{center}
\fbox{\begin{minipage}{0.7\textwidth}
\vspace{1em}
\begin{center}
{\Large\bfseries Appendix Table of Contents}
\end{center}
\vspace{0.5em}
\etocsettagdepth{chapter}{none}
\etocsettagdepth{main}{none}
\etocsettagdepth{appendix}{section}
\etocsettocstyle{}{}
\tableofcontents
\vspace{1em}
\end{minipage}}
\end{center}
\vspace{2em}

\section{Effective Rank of Weight Matrices}
\label{sec:effective-rank}

To characterize the intrinsic dimensionality of neural network weight matrices, we adopt the notion of \emph{effective rank}~\citep{roy2007effective}, which provides a continuous measure of how many singular components contribute meaningfully to a matrix's structure.

\subsection{Definition}
\label{subsec:erank-definition}

Let $W \in \mathbb{R}^{m \times n}$ denote a weight matrix, and let $Q = \min(m, n)$. The singular value decomposition (SVD) of $W$ yields singular values
\begin{equation}
\label{eq:singular-values}
\sigma_1(W) \geq \sigma_2(W) \geq \cdots \geq \sigma_Q(W) \geq 0.
\end{equation}
We define the \emph{normalized singular value distribution} via $\ell_1$ normalization:
\begin{equation}
\label{eq:normalized-sv}
p_i(W) = \frac{\sigma_i(W)}{\sum_{j=1}^{Q} \sigma_j(W)}, \quad i = 1, \ldots, Q.
\end{equation}
The Shannon entropy of this distribution is given by
\begin{equation}
\label{eq:sv-entropy}
H(W) = -\sum_{i=1}^{Q} p_i(W) \log p_i(W),
\end{equation}
where we adopt the convention that $0 \log 0 = 0$. The \emph{effective rank} is then defined as
\begin{equation}
\label{eq:erank-def}
\rho_{\text{eff}}(W) = \exp\bigl(H(W)\bigr) \in [1, Q].
\end{equation}

The effective rank admits an intuitive interpretation. When a single singular value dominates the spectrum (i.e., $p_1 \approx 1$), the entropy approaches zero and $\rho_{\text{eff}}(W) \approx 1$, indicating that the matrix is approximately rank-one. Conversely, when all nonzero singular values are equal, the distribution is uniform and $\rho_{\text{eff}}(W)$ equals the numerical rank of $W$. Thus, the effective rank provides a smooth interpolation between these extremes, quantifying how the matrix's energy is distributed across its singular components.

\paragraph{Aggregation Across Layers.}
To obtain a single predictor for multi-matrix compression, we aggregate effective ranks across weight matrices $\{W_i\}_{i=1}^{L}$ using a parameter-weighted mean:
\begin{equation}
\label{eq:aggregate-effective-rank}
\bar{\rho}_{\text{eff}} = \frac{\sum_{i=1}^{L} n_i \cdot \rho_{\text{eff}}(W_i)}{\sum_{i=1}^{L} n_i}, \quad n_i = \mathrm{rows}(W_i) \times \mathrm{cols}(W_i).
\end{equation}
This aggregation mirrors that of stable rank (\Cref{eq:aggregate-stable-rank}) and was chosen empirically for yielding strong correlations with degradation metrics.

\subsection{Application to Low-Rank Approximation}
\label{subsec:erank-truncation}

The effective rank serves as a natural guide for selecting the truncation rank in low-rank matrix approximations. Let $W_k$ denote the optimal rank-$k$ approximation obtained via truncated SVD:
\begin{equation}
\label{eq:truncated-svd}
W_k = \sum_{i=1}^{k} \sigma_i(W) \, u_i v_i^\top,
\end{equation}
where $u_i$ and $v_i$ are the left and right singular vectors corresponding to $\sigma_i(W)$. The effective rank provides a \emph{spectrum-aware reference} for the truncation rank:
\begin{equation}
\label{eq:k-reference}
k_{\mathrm{ref}} = \lceil \rho_{\text{eff}}(W) \rceil,
\end{equation}
which estimates the number of singular directions that are effectively active in $W$.

\paragraph{Relationship to approximation error.}
The quality of the truncated approximation is typically measured by the relative Frobenius norm error:
\begin{equation}
\label{eq:frobenius-error}
\frac{\|W - W_k\|_F^2}{\|W\|_F^2} = 1 - \frac{\sum_{i=1}^{k} \sigma_i(W)^2}{\sum_{i=1}^{Q} \sigma_i(W)^2}.
\end{equation}
In practice, the truncation rank $k$ is often selected to retain a target fraction of the total spectral energy (e.g., 95\% or 99\%). The effective rank complements this approach by providing an \emph{a priori} estimate of the matrix's compressibility: matrices with small effective rank relative to their ambient dimension are more amenable to aggressive low-rank approximation with minimal reconstruction error.

\section{Reproducibility}
\label{sec:reproducibility}

\subsection{Models and Data}

\paragraph{Model Families.}
We evaluate compression on two publicly available model families:
\begin{itemize}[nosep]
    \item \textbf{Qwen3}: 0.6B, 1.7B, 4B, 8B, 14B parameter variants
    \item \textbf{Gemma3}: 270M, 1B, 4B, 12B, 27B parameter variants (instruction-tuned)
\end{itemize}
All models are accessed via HuggingFace Transformers library.

\paragraph{Evaluation Benchmarks.}
We evaluate on eight benchmarks using the LM Evaluation Harness \citep{eval-harness}:
WikiText-2 (perplexity), ARC-Challenge, ARC-E, BoolQ, HellaSwag, PIQA, and WinoGrande (accuracy).
Default evaluation settings are used for all benchmarks.

\subsection{Compression Configuration}

\paragraph{SVD Truncation.}
We apply truncated SVD to weight matrices without compensation mechanisms to isolate raw architectural responses to compression. Truncation ranks are varied systematically to achieve compression ratios $\gamma \in [0.1, 0.9]$.

\paragraph{Layer Configurations.}
Three compression scenarios are evaluated:
\begin{itemize}[nosep]
    \item \textbf{ATTN-only}: Compress $W_Q, W_K, W_V, W_O$ matrices
    \item \textbf{MLP-only}: Compress $W_{\text{gate}}, W_{\text{up}}, W_{\text{down}}$ matrices
    \item \textbf{Combined}: Compress both attention and MLP layers
\end{itemize}

\subsection{Compute Resources}

All experiments were conducted on NVIDIA H200 GPUs (141GB memory) with 16 vCPUs.

\subsection{Statistical Analysis}

\paragraph{Symbolic Regression.}
We use \texttt{gplearn} for symbolic regression with tournament selection and genetic operators. A parsimony coefficient penalizes overly complex formulas.

\paragraph{Evaluation Protocol.}
All reported correlations use leave-one-out cross-validation (LOO-CV) to assess generalization. Training correlations are reported alongside LOO correlations to identify overfitting.

\paragraph{Sample Sizes.}
Each layer-type configuration includes data points across model scales and compression ratios within each model family. \Cref{tab:method-overview} provides the complete breakdown by compression method and layer type.

\begin{table}[H]
\caption{Sample sizes by compression method and layer type.}
\label{tab:method-overview}
\centering
\small
\begin{tabular}{@{}lcccc@{}}
\toprule
Method & ATTN & MLP & BOTH & Total \\
\midrule
Vanilla SVD & 26 & 18 & 16 & 60 \\
ASVD & 16 & 16 & 16 & 48 \\
ASVD (stable ranks) & 16 & 16 & 16 & 48 \\
SVD-LLM & 16 & 16 & 16 & 48 \\
\midrule
\textbf{Total} & \textbf{74} & \textbf{66} & \textbf{64} & \textbf{204} \\
\bottomrule
\end{tabular}
\end{table}

\subsection{Code Availability}

The source code for linear regression with proposed template formulas and those discovered by symbolic regression, is in the supplementary materials.

The rest of the code for reproducing our experiments, including compression scripts, evaluation pipelines, and symbolic regression analysis, will be made available upon publication.

\section{Limitations}
\label{sec:limitations}

\paragraph{Model Family Coverage.}
Our experiments focus on Qwen3 and Gemma3 model families, comprising 16--26 samples per method per layer configuration (\Cref{tab:method-overview}). Generalization to other architectures (e.g., LLaMA, Mistral) is untested. Cross-family validation (training on one family, testing on another) has not been systematically performed.

\paragraph{Statistical Validation.}
Bootstrap confidence intervals for LOO correlation estimates have not been computed. Baseline comparisons against simple functional forms (e.g., $\Delta = a\gamma + b$) to verify that discovered formulas provide genuine improvement over trivial alternatives are not yet included.

\paragraph{Functional Form Ambiguity.}
Different tasks favor different functional forms: $\gamma \cdot \bar{\rho}_{\text{eff}}$ dominates for ARC-Challenge, ARC-Easy, and WinoGrande, $\gamma \cdot \bar{\rho}_s$ performs best for BoolQ, $\log(\bar{\rho}_s) + \log N$ leads for HellaSwag and PIQA, and $\mathcal{B} + \bar{\rho}_s$ works best for WikiText perplexity. Whether this variation reflects genuine task-specific structure or fitting noise cannot be determined without additional validation.

\paragraph{Theoretical Explanations.}
The proposed connection between operation type (tensor contractions vs. Hadamard products) and predictability is a hypothesis supported by correlational evidence. Controlled experiments isolating this mechanism have not been performed.

\section{Broader Impact Statement}
\label{sec:broader-impact}

This work identifies predictive formulas for compression-induced degradation in large language models, enabling practitioners to estimate performance loss before committing to expensive evaluation. We discuss potential impacts below.

\paragraph{Positive Impacts.}
Our findings enable more predictable deployment of compressed language models by identifying when compression effects can be reliably anticipated (in attention layers) and when task-specific calibration is required (in MLP layers). This predictability reduces the risk of deploying models with unexpectedly degraded performance, potentially improving the reliability of AI systems in production. Furthermore, by enabling practitioners to estimate degradation from weight spectra before committing to expensive compression-evaluation loops, our predict-then-compress workflow reduces the computational resources wasted on unpromising configurations, lowering the energy consumption and carbon emissions associated with compression research and development.

\paragraph{Potential Negative Impacts.}
By making compression outcomes more predictable, this work could accelerate the deployment of compressed language models, inheriting any risks associated with LLM misuse. However, we note that: (1) our work analyzes existing compression techniques rather than introducing new capabilities; (2) the models studied (Qwen3, Gemma3) are already publicly available; and (3) compression primarily affects inference efficiency rather than model capabilities. The predictive formulas we identify do not enable new forms of harm beyond what is already possible with existing compressed models.

\paragraph{Limitations of Impact Assessment.}
Our analysis focuses on SVD-based compression methods, including both uncompensated (vanilla SVD) and activation-aware variants (ASVD, SVD-LLM). Other compression paradigms, such as quantization and pruning may exhibit different scaling behaviors. Additionally, our experiments are limited to two model families; generalization of both the technical findings and their societal implications to other architectures remains to be validated.

\section{Formula Template Catalog}
\label{sec:formula-templates}

This appendix provides the complete catalog of 42 formula templates introduced in \Cref{subsubsec:formula-templates-intro}. Each template is presented with its explicit functional form and design rationale.

\subsection{Single-Variable Templates}

These templates test whether a single predictor suffices to explain performance variation under compression.

\paragraph{Linear Compression (F1).}
The most basic hypothesis posits that performance degrades linearly with compression ratio:
\begin{equation}
\label{eq:F1}
y = \alpha_0 + \alpha_1 \gamma
\end{equation}
where $\gamma = N_{\text{comp}}/N$ is the fraction of parameters retained.

\paragraph{Log-Compression (F2).}
Logarithmic transformation captures diminishing returns of compression:
\begin{equation}
\label{eq:F2}
y = \alpha_0 + \alpha_1 \log \gamma
\end{equation}
This formulation is equivalent to modeling performance as a function of $\log N - \log N_{\text{comp}}$.

\paragraph{Density-Compression Ratio (F3).}
Information density relative to compression level:
\begin{equation}
\label{eq:F3}
y = \alpha_0 + \alpha_1 \frac{\mathcal{B}}{\gamma}
\end{equation}
where $\mathcal{B}$ denotes bits per parameter under entropy coding.

\paragraph{Scale Difference (F4).}
The difference in log-scale between original and compressed models:
\begin{equation}
\label{eq:F4}
y = \alpha_0 + \alpha_1 (\log N - \log N_{\text{comp}})
\end{equation}
Note that $\log N - \log N_{\text{comp}} = -\log \gamma$, making this equivalent to F2.

\subsection{Two-Variable Templates}

These templates test whether combining two predictors improves upon single-variable models.

\paragraph{Compression-Scale (F5).}
Joint effect of compression ratio and model scale:
\begin{equation}
\label{eq:F5}
y = \alpha_0 + \alpha_1 \gamma + \alpha_2 \log N
\end{equation}

\paragraph{Compression-Density (F6).}
Compression ratio combined with information density:
\begin{equation}
\label{eq:F6}
y = \alpha_0 + \alpha_1 \gamma + \alpha_2 \mathcal{B}
\end{equation}

\paragraph{Dual-Scale (F7).}
Original and compressed scale as independent predictors:
\begin{equation}
\label{eq:F7}
y = \alpha_0 + \alpha_1 \log N + \alpha_2 \log N_{\text{comp}}
\end{equation}
This separates the effects of original model capacity from retained capacity.

\paragraph{Density-Rank (F8).}
Information density combined with spectral properties:
\begin{equation}
\label{eq:F8}
y = \alpha_0 + \alpha_1 \mathcal{B} + \alpha_2 \bar{\rho}_s
\end{equation}
where $\bar{\rho}_s$ is the mean stable rank across weight matrices.

\paragraph{Ratio-Difference (F9).}
Density ratio with log-scale difference:
\begin{equation}
\label{eq:F9}
y = \alpha_0 + \alpha_1 \frac{\mathcal{B}}{\gamma} + \alpha_2 (\log N - \log N_{\text{comp}})
\end{equation}

\paragraph{Nonlinear Compression (F10).}
Quadratic and exponential transformations of compression ratio:
\begin{equation}
\label{eq:F10}
y = \alpha_0 + \alpha_1 \gamma^2 + \alpha_2 e^{-\gamma}
\end{equation}
This captures nonlinear saturation effects at extreme compression levels.

\subsection{Three-Variable Templates}

The maximum complexity allowed under our parsimony constraints, these templates combine three predictors.

\paragraph{Compression-Scale-Density (F11).}
Full linear model with compression, scale, and density:
\begin{equation}
\label{eq:F11}
y = \alpha_0 + \alpha_1 \gamma + \alpha_2 \log N + \alpha_3 \mathcal{B}
\end{equation}

\paragraph{Compression-CompressedScale-Density (F12).}
Replacing original scale with compressed scale:
\begin{equation}
\label{eq:F12}
y = \alpha_0 + \alpha_1 \gamma + \alpha_2 \log N_{\text{comp}} + \alpha_3 \mathcal{B}
\end{equation}

\paragraph{Dual-Scale-Density (F13).}
Both scale measures with density:
\begin{equation}
\label{eq:F13}
y = \alpha_0 + \alpha_1 \log N + \alpha_2 \log N_{\text{comp}} + \alpha_3 \mathcal{B}
\end{equation}

\paragraph{Compression-Rank-Density (F14).}
Compression ratio with spectral and density terms:
\begin{equation}
\label{eq:F14}
y = \alpha_0 + \alpha_1 \gamma + \alpha_2 \bar{\rho}_s + \alpha_3 \mathcal{B}
\end{equation}

\paragraph{Nonlinear-Scale (F15).}
Nonlinear compression terms with compressed scale:
\begin{equation}
\label{eq:F15}
y = \alpha_0 + \alpha_1 \gamma^2 + \alpha_2 e^{-\gamma} + \alpha_3 \log N_{\text{comp}}
\end{equation}
This extends F10 by incorporating model scale information.

\subsection{Interaction Templates}

These templates test multiplicative relationships between predictors, encoding hypotheses about synergistic effects.

\paragraph{Compression-EffectiveRank Interaction (F16).}
Product of compression ratio and effective rank:
\begin{equation}
\label{eq:F16}
y = \alpha_0 + \alpha_1 (\gamma \cdot \bar{\rho}_{\text{eff}})
\end{equation}
where $\bar{\rho}_{\text{eff}}$ is the mean effective rank. This tests whether compression impact scales with the intrinsic dimensionality of the weight matrices.

\paragraph{Density-StableRank Interaction (F17).}
Product of information density and stable rank:
\begin{equation}
\label{eq:F17}
y = \alpha_0 + \alpha_1 (\mathcal{B} \cdot \bar{\rho}_s)
\end{equation}

\paragraph{Compression-Density Interaction (F18).}
Full interaction model with main effects:
\begin{equation}
\label{eq:F18}
y = \alpha_0 + \alpha_1 \gamma + \alpha_2 \mathcal{B} + \alpha_3 (\gamma \cdot \mathcal{B})
\end{equation}
This tests whether the effect of compression depends on information density, and vice versa.

\subsection{Template Summary}

\Cref{tab:template-summary} provides a compact reference for all 42 templates.

\begin{table}[htbp]
\caption{Summary of formula templates by category and variable count.}
\label{tab:template-summary}
\centering
\small
\resizebox{0.7\linewidth}{!}{%
\begin{tabular}{@{}llcll@{}}
\toprule
ID & Name & Vars & Formula Structure & Scope \\
\midrule
\multicolumn{5}{@{}l}{\textit{Single-Variable (4 templates)}} \\
F1 & Linear compression & 1 & $\gamma$ & All \\
F2 & Log-compression & 1 & $\log \gamma$ & All \\
F3 & Density ratio & 1 & $\mathcal{B}/\gamma$ & All \\
F4 & Scale difference & 1 & $\log N - \log N_{\text{comp}}$ & All \\
\midrule
\multicolumn{5}{@{}l}{\textit{Two-Variable (6 templates)}} \\
F5 & Compression-scale & 2 & $\gamma, \log N$ & All \\
F6 & Compression-density & 2 & $\gamma, \mathcal{B}$ & All \\
F7 & Dual-scale & 2 & $\log N, \log N_{\text{comp}}$ & All \\
F8 & Density-rank & 2 & $\mathcal{B}, \bar{\rho}_s$ & All \\
F9 & Ratio-difference & 2 & $\mathcal{B}/\gamma, \log N - \log N_{\text{comp}}$ & All \\
F10 & Nonlinear compression & 2 & $\gamma^2, e^{-\gamma}$ & All \\
\midrule
\multicolumn{5}{@{}l}{\textit{Three-Variable (5 templates)}} \\
F11 & Compression-scale-density & 3 & $\gamma, \log N, \mathcal{B}$ & All \\
F12 & Compression-compScale-density & 3 & $\gamma, \log N_{\text{comp}}, \mathcal{B}$ & All \\
F13 & Dual-scale-density & 3 & $\log N, \log N_{\text{comp}}, \mathcal{B}$ & All \\
F14 & Compression-rank-density & 3 & $\gamma, \bar{\rho}_s, \mathcal{B}$ & All \\
F15 & Nonlinear-scale & 3 & $\gamma^2, e^{-\gamma}, \log N_{\text{comp}}$ & All \\
\midrule
\multicolumn{5}{@{}l}{\textit{Interaction (3 templates)}} \\
F16 & Compression-effRank & 1 & $\gamma \cdot \bar{\rho}_{\text{eff}}$ & All \\
F17 & Density-stableRank & 1 & $\mathcal{B} \cdot \bar{\rho}_s$ & All \\
F18 & Compression-density interaction & 3 & $\gamma, \mathcal{B}, \gamma \cdot \mathcal{B}$ & All \\
\midrule
\multicolumn{5}{@{}l}{\textit{Threshold-Corrected (4 templates)}} \\
F19 & Inverse threshold & 3 & $\gamma^2, 1/\gamma - 1, \log N_{\text{comp}}$ & All \\
F20 & Simplified inverse & 3 & $\gamma^2, 1/\gamma, \log N_{\text{comp}}$ & All \\
F21 & Logarithmic threshold & 3 & $\gamma^2, \log(1/\gamma), \log N_{\text{comp}}$ & All \\
F22 & Exponential inverse & 3 & $\gamma^2, e^{1/\gamma-1}, \log N_{\text{comp}}$ & All \\
\midrule
\multicolumn{5}{@{}l}{\textit{Entropy-Based (4 templates)}} \\
F23 & Entropy-compression & 2 & $\gamma, H$ & $\mathrm{P}$ \\
F24 & Entropy-scale & 2 & $H, \log N$ & $\mathrm{P}$ \\
F25 & Entropy-compression interaction & 3 & $\gamma, H, \gamma \cdot H$ & $\mathrm{P}$ \\
F26 & Entropy-density & 2 & $H, \mathcal{B}$ & $\mathrm{P}$ \\
\midrule
\multicolumn{5}{@{}l}{\textit{Layer-Specific (3 templates)}} \\
F27 & Dual-layer compression & 2 & $\gamma_{\text{attn}}, \gamma_{\text{mlp}}$ & All \\
F28 & Layer-weighted & 3 & $\gamma_{\text{attn}}, \gamma_{\text{mlp}}, \log N$ & All \\
F29 & Layer ratio & 1 & $\gamma_{\text{attn}}/\gamma_{\text{mlp}}$ & All \\
\midrule
\multicolumn{5}{@{}l}{\textit{Rank-Based (3 templates)}} \\
F30 & Direct rank & 1 & $\log r$ & All \\
F31 & Rank-scale & 2 & $\log r, \log N$ & All \\
F32 & Rank-density & 2 & $r/\bar{\rho}_s, \mathcal{B}$ & All \\
\midrule
\multicolumn{5}{@{}l}{\textit{Energy-Based (2 templates)}} \\
F33 & Energy retention & 2 & $\bar{k}_{95}, \gamma$ & All \\
F34 & Energy gap & 2 & $\bar{k}_{99} - \bar{k}_{95}, \log N_{\text{comp}}$ & All \\
\midrule
\multicolumn{5}{@{}l}{\textit{Baseline-Normalized (8 templates)}} \\
F35 & Relative $\mathrm{P}$ degradation & 2 & $\gamma, \log N$ & $\mathrm{P}$ \\
F36 & Log-$\mathrm{P}$ ratio & 2 & $\gamma, \log N_{\text{comp}}$ & $\mathrm{P}$ \\
F37 & Log-$\mathrm{P}$ with baseline & 2 & $\gamma, \log(\mathrm{P}_0)$ & $\mathrm{P}$ \\
F38 & Log-$\mathrm{P}$ with baseline+scale & 3 & $\gamma, \log(\mathrm{P}_0), \log N$ & $\mathrm{P}$ \\
F39 & Accuracy drop & 2 & $\gamma, \log N$ & ACC \\
F40 & Relative accuracy degradation & 2 & $\gamma, \log N_{\text{comp}}$ & ACC \\
F41 & Accuracy with baseline & 3 & $\gamma, \mathcal{A}_0, \log N$ & ACC \\
F42 & Log-$\mathrm{P}$ ratio with entropy & 2 & $\gamma, H$ & $\mathrm{P}$ \\
\bottomrule
\end{tabular}%
}
\end{table}

\paragraph{Summary.}
The 42 templates span single-variable through three-variable forms, testing compression ratio, model scale, spectral properties, and their interactions. This systematic coverage enables identification of dominant predictors through cross-validation comparison, as detailed in \Cref{sec:compare}.

\subsection{Threshold-Corrected Templates}

The exponential decay term $e^{-\gamma}$ in templates such as F10 and F15 exhibits problematic behavior at extreme compression: as $\gamma \to 0$, the term remains bounded ($e^{-\gamma} \to 1$), failing to capture the severe performance degradation expected under aggressive compression. We propose four alternative templates that incorporate threshold terms with correct asymptotic behavior-diverging as $\gamma \to 0$ and vanishing as $\gamma \to 1$.

\paragraph{Inverse Threshold (F19).}
Replacing exponential decay with an inverse threshold term:
\begin{equation}
\label{eq:F19}
y = \alpha_0 + \alpha_1 \gamma^2 + \alpha_2 \left(\frac{1}{\gamma} - 1\right) + \alpha_3 \log N_{\text{comp}}
\end{equation}
The term $(1/\gamma - 1)$ provides the desired threshold behavior: it diverges as $\gamma \to 0$ (extreme compression) and vanishes at $\gamma = 1$ (no compression). The offset ensures zero penalty at full capacity, improving interpretability.

\paragraph{Simplified Inverse Threshold (F20).}
A reduced form absorbing the constant offset into the intercept:
\begin{equation}
\label{eq:F20}
y = \alpha_0 + \alpha_1 \gamma^2 + \frac{\alpha_2}{\gamma} + \alpha_3 \log N_{\text{comp}}
\end{equation}
This simplification maintains equivalent asymptotic behavior while reducing algebraic complexity. The inverse relationship encodes ``penalty per unit of retained capacity.''

\paragraph{Logarithmic Threshold (F21).}
A softer divergence using logarithmic transformation:
\begin{equation}
\label{eq:F21}
y = \alpha_0 + \alpha_1 \gamma^2 + \alpha_2 \log\left(\frac{1}{\gamma}\right) + \alpha_3 \log N_{\text{comp}}
\end{equation}
Since $\log(1/\gamma) = -\log \gamma$, this template captures threshold effects with logarithmic rather than polynomial divergence. The slower growth rate may provide better generalization when extreme compression points are sparse in the training data.

\paragraph{Exponential Inverse Threshold (F22).}
The strongest threshold effect, combining exponential and inverse transformations:
\begin{equation}
\label{eq:F22}
y = \alpha_0 + \alpha_1 \gamma^2 + \alpha_2 \exp\left(\frac{1}{\gamma} - 1\right) + \alpha_3 \log N_{\text{comp}}
\end{equation}
This formulation produces the fastest divergence as $\gamma \to 0$, potentially capturing catastrophic failure modes under severe compression. However, the aggressive nonlinearity increases overfitting risk.

\paragraph{Threshold Behavior Comparison.}
\Cref{tab:threshold-comparison} contrasts the asymptotic behavior of the original and proposed threshold terms.

\begin{table}[H]
\caption{Asymptotic behavior of threshold terms under extreme and minimal compression.}
\label{tab:threshold-comparison}
\centering
\small
\begin{tabular}{@{}lcccl@{}}
\toprule
Term & $\gamma \to 0$ & $\gamma = 0.5$ & $\gamma \to 1$ & Divergence \\
\midrule
$e^{-\gamma}$ (F10, F15) & $1$ & $0.61$ & $0.37$ & Bounded \\
$1/\gamma - 1$ (F19) & $\infty$ & $1$ & $0$ & Linear \\
$1/\gamma$ (F20) & $\infty$ & $2$ & $1$ & Linear \\
$\log(1/\gamma)$ (F21) & $\infty$ & $0.69$ & $0$ & Logarithmic \\
$e^{1/\gamma - 1}$ (F22) & $\infty$ & $2.72$ & $1$ & Exponential \\
\bottomrule
\end{tabular}
\end{table}

\paragraph{Design Rationale.}
These threshold-corrected templates address a fundamental limitation in existing formulations: the bounded nature of $e^{-\gamma}$ prevents the model from capturing threshold effects where performance degrades catastrophically below a critical compression level. By introducing terms that diverge as $\gamma \to 0$, we enable the regression to fit cliff-like degradation curves commonly observed in practice. The hierarchy from F19 to F22 provides a spectrum of divergence rates, allowing empirical selection based on cross-validation performance.

\subsection{Entropy-Based Templates (Perplexity Only)}

These templates incorporate dataset entropy $H$, measured from embedding layer activations. Since entropy directly relates to token prediction uncertainty, these templates are \textbf{only applicable to perplexity ($\mathrm{P}$) prediction}, not accuracy tasks.

\paragraph{Entropy-Compression (F23).}
Combining compression ratio with dataset entropy:
\begin{equation}
\label{eq:F23}
\mathrm{P} = \alpha_0 + \alpha_1 \gamma + \alpha_2 H
\end{equation}
where $H$ denotes dataset entropy computed from embedding activations. This tests whether data complexity modulates compression sensitivity.

\paragraph{Entropy-Scale (F24).}
Dataset entropy combined with model scale:
\begin{equation}
\label{eq:F24}
\mathrm{P} = \alpha_0 + \alpha_1 H + \alpha_2 \log N
\end{equation}
This separates the effects of data complexity from model capacity.

\paragraph{Entropy-Compression Interaction (F25).}
Full interaction model testing whether compression impact depends on data entropy:
\begin{equation}
\label{eq:F25}
\mathrm{P} = \alpha_0 + \alpha_1 \gamma + \alpha_2 H + \alpha_3 (\gamma \cdot H)
\end{equation}
The interaction term tests whether higher-entropy data requires more parameters to maintain performance under compression.

\paragraph{Entropy-Density (F26).}
Two information-theoretic measures combined:
\begin{equation}
\label{eq:F26}
\mathrm{P} = \alpha_0 + \alpha_1 H + \alpha_2 \mathcal{B}
\end{equation}
Both dataset entropy and bits-per-parameter measure information content, testing their joint predictive power.

\subsection{Layer-Specific Templates}

These templates use separate compression ratios for attention ($\gamma_{\text{attn}}$) and MLP ($\gamma_{\text{mlp}}$) layers, testing whether different layer types exhibit different compression sensitivity.

\paragraph{Dual-Layer Compression (F27).}
Independent compression effects for each layer type:
\begin{equation}
\label{eq:F27}
y = \alpha_0 + \alpha_1 \gamma_{\text{attn}} + \alpha_2 \gamma_{\text{mlp}}
\end{equation}
This tests whether attention and MLP layers contribute differently to performance under compression.

\paragraph{Layer-Weighted Compression (F28).}
Layer-specific compression with scale control:
\begin{equation}
\label{eq:F28}
y = \alpha_0 + \alpha_1 \gamma_{\text{attn}} + \alpha_2 \gamma_{\text{mlp}} + \alpha_3 \log N
\end{equation}
Separates layer-specific effects while controlling for overall model capacity.

\paragraph{Layer Ratio (F29).}
Relative compression between layer types:
\begin{equation}
\label{eq:F29}
y = \alpha_0 + \alpha_1 \frac{\gamma_{\text{attn}}}{\gamma_{\text{mlp}}}
\end{equation}
Tests whether the \emph{balance} of compression between layers matters more than absolute compression levels.

\subsection{Rank-Based Templates}

These templates use the SVD truncation rank $r$ directly, rather than the derived compression ratio. This provides a more direct measure of retained spectral information.

\paragraph{Direct Rank (F30).}
Log-transformed truncation rank as sole predictor:
\begin{equation}
\label{eq:F30}
y = \alpha_0 + \alpha_1 \log r
\end{equation}
The logarithmic transformation captures diminishing returns from increasing rank.

\paragraph{Rank-Scale (F31).}
Truncation rank with model scale:
\begin{equation}
\label{eq:F31}
y = \alpha_0 + \alpha_1 \log r + \alpha_2 \log N
\end{equation}
Tests whether optimal rank scales with model size.

\paragraph{Rank-Density (F32).}
Ratio of truncation rank to stable rank:
\begin{equation}
\label{eq:F32}
y = \alpha_0 + \alpha_1 \frac{r}{\bar{\rho}_s} + \alpha_2 \mathcal{B}
\end{equation}
The ratio $r/\bar{\rho}_s$ indicates what fraction of the ``useful'' spectral content is retained by the truncation.

\subsection{Energy-Based Templates}

These templates use spectral energy metrics $\bar{k}_{95}$ and $\bar{k}_{99}$, representing the mean rank required to capture 95\% and 99\% of spectral energy across weight matrices.

\paragraph{Energy Retention (F33).}
Spectral concentration combined with compression:
\begin{equation}
\label{eq:F33}
y = \alpha_0 + \alpha_1 \bar{k}_{95} + \alpha_2 \gamma
\end{equation}
Models with flatter spectra (higher $\bar{k}_{95}$) may be more sensitive to compression since information is spread across more dimensions.

\paragraph{Energy Gap (F34).}
Spectral tail heaviness:
\begin{equation}
\label{eq:F34}
y = \alpha_0 + \alpha_1 (\bar{k}_{99} - \bar{k}_{95}) + \alpha_2 \log N_{\text{comp}}
\end{equation}
The gap $\bar{k}_{99} - \bar{k}_{95}$ measures how much additional rank is needed to capture the last 4\% of spectral energy, indicating tail heaviness.

\subsection{Baseline-Normalized Templates}

These templates incorporate baseline (uncompressed) performance metrics to normalize or predict degradation. Using relative metrics removes model-specific offsets, focusing purely on the compression-induced change.

\paragraph{Relative $\mathrm{P}$ Degradation (F35).}
Perplexity increase normalized by baseline:
\begin{equation}
\label{eq:F35}
\frac{\mathrm{P} - \mathrm{P}_0}{\mathrm{P}_0} = \alpha_0 + \alpha_1 \gamma + \alpha_2 \log N
\end{equation}
Normalizing by baseline $\mathrm{P}$ focuses on relative degradation, removing model-specific baseline differences and enabling fair comparison across model families.

\paragraph{Log-$\mathrm{P}$ Ratio (F36).}
Multiplicative perplexity increase in log space:
\begin{equation}
\label{eq:F36}
\log\left(\frac{\mathrm{P}}{\mathrm{P}_0}\right) = \alpha_0 + \alpha_1 \gamma + \alpha_2 \log N_{\text{comp}}
\end{equation}
The log-ratio measures the multiplicative increase in perplexity due to compression, providing a scale-invariant degradation metric.

\paragraph{Log-$\mathrm{P}$ with Baseline (F37).}
Baseline log-perplexity as a covariate:
\begin{equation}
\label{eq:F37}
\log(\mathrm{P}) = \alpha_0 + \alpha_1 \gamma + \alpha_2 \log(\mathrm{P}_0)
\end{equation}
Including baseline log-$\mathrm{P}$ as a covariate captures model quality; the compression effect is additive in log-space.

\paragraph{Log-$\mathrm{P}$ with Baseline and Scale (F38).}
Full model incorporating compression, baseline quality, and scale:
\begin{equation}
\label{eq:F38}
\log(\mathrm{P}) = \alpha_0 + \alpha_1 \gamma + \alpha_2 \log(\mathrm{P}_0) + \alpha_3 \log N
\end{equation}
This comprehensive template separates the contributions of compression ratio, inherent model quality, and model capacity.

\paragraph{Accuracy Drop (F39).}
Absolute accuracy degradation:
\begin{equation}
\label{eq:F39}
\mathcal{A}_0 - \mathcal{A} = \alpha_0 + \alpha_1 \gamma + \alpha_2 \log N
\end{equation}
Direct prediction of accuracy loss as a function of compression and scale, applicable to any accuracy-based benchmark.

\paragraph{Relative Accuracy Degradation (F40).}
Accuracy loss normalized by baseline performance:
\begin{equation}
\label{eq:F40}
\frac{\mathcal{A}_0 - \mathcal{A}}{\mathcal{A}_0} = \alpha_0 + \alpha_1 \gamma + \alpha_2 \log N_{\text{comp}}
\end{equation}
Relative accuracy loss enables comparison across tasks with different baseline difficulty levels.

\paragraph{Accuracy with Baseline Covariate (F41).}
Predicting compressed accuracy from baseline and compression parameters:
\begin{equation}
\label{eq:F41}
\mathcal{A} = \alpha_0 + \alpha_1 \gamma + \alpha_2 \mathcal{A}_0 + \alpha_3 \log N
\end{equation}
Using baseline accuracy as a covariate allows direct prediction of post-compression accuracy.

\paragraph{Log-$\mathrm{P}$ Ratio with Entropy (F42).}
Testing entropy modulation of relative perplexity increase:
\begin{equation}
\label{eq:F42}
\log\left(\frac{\mathrm{P}}{\mathrm{P}_0}\right) = \alpha_0 + \alpha_1 \gamma + \alpha_2 H
\end{equation}
This template tests whether dataset entropy modulates the relative $\mathrm{P}$ increase under compression, combining the information-theoretic perspective with baseline normalization.

\section{Symbolic Regression Discoveries}
\label{sec:gplearn-formulas}

In addition to the manually designed templates (\Cref{sec:formula-templates}), we employ genetic programming via \texttt{gplearn} to search the space of symbolic expressions for predictive formulas. This data-driven approach complements our hypothesis-driven templates by potentially discovering unexpected functional relationships. We report 20 unique formula structures that satisfy our parsimony constraints ($\leq 3$ variables, $\leq 2$ nonlinear operations) and achieve non-trivial correlation with compression-induced degradation. Formulas are presented in descending order of training correlation.

\subsection{High-Correlation Discoveries (D1--D7)}

These formulas achieved training correlation $R > 0.56$, representing the most predictive structures discovered.

\paragraph{Log-Stable-Rank over Log-Compressed-Size (D1).}
The ratio of log-stable-rank to log-compressed-model-size:
\begin{equation}
\label{eq:D1}
y = \frac{\log \bar{\rho}_s}{\log N_{\text{comp}}}
\end{equation}
This formulation achieved Train $R = 0.607$ and LOO $R = 0.510$ on HellaSwag (both layers), suggesting that the normalized rank complexity-stable rank relative to model scale-serves as a robust predictor of compression tolerance.

\paragraph{Bits per Compression Ratio (D2).}
Information density scaled by compression level:
\begin{equation}
\label{eq:D2}
y = \frac{\mathcal{B}}{\gamma}
\end{equation}
This formula achieved Train $R = 0.600$ and LOO $R = 0.504$ on attention layer experiments. Discovered independently across multiple configurations, it encodes the hypothesis that degradation scales with ``bits per unit of retained capacity.''

\paragraph{Log of Rank-Entropy Ratio (D3).}
Logarithm incorporating stable rank, truncation rank, and entropy:
\begin{equation}
\label{eq:D3}
y = \log\left((\bar{\rho}_s + r) \cdot \frac{c}{H}\right)
\end{equation}
where $r$ denotes the SVD truncation rank and $c$ is a constant. This three-variable formula achieved Train $R = 0.599$ and LOO $R = 0.479$ on WikiText (both layers).

\paragraph{Gamma-Entropy-Exponential Sum (D4).}
A multi-term formula combining linear and exponential components:
\begin{equation}
\label{eq:D4}
y = \gamma + H + e^{\gamma} + c
\end{equation}
This formula achieved Train $R = 0.590$ and LOO $R = 0.472$ on WikiText (both layers). Notably, it includes $e^{\gamma}$ rather than $e^{-\gamma}$, capturing scenarios where performance improves exponentially with retained capacity.

\paragraph{Log Stable-Rank over Shifted Entropy (D5).}
Logarithm of the ratio between stable rank and shifted entropy:
\begin{equation}
\label{eq:D5}
y = \log\left(\frac{\bar{\rho}_s}{H + c}\right)
\end{equation}
This formulation achieved Train $R = 0.585$ and LOO $R = 0.468$ on WikiText (MLP layer), connecting spectral properties of weight matrices to data complexity through a log-ratio structure.

\paragraph{Inverse of Shifted Bits (D6).}
Inverse density with a shift parameter:
\begin{equation}
\label{eq:D6}
y = \frac{1}{\mathcal{B} + c}
\end{equation}
This formula achieved Train $R = 0.565$ and LOO $R = 0.452$ on HellaSwag (MLP layer). The shift term prevents singularity at low bit-rates and captures saturation effects.

\paragraph{Simple Inverse of Bits (D7).}
The most parsimonious density-based predictor:
\begin{equation}
\label{eq:D7}
y = \frac{1}{\mathcal{B}}
\end{equation}
Despite its simplicity, this single-variable formula achieved Train $R = 0.564$ and the best LOO correlation ($R = 0.525$) among all constrained discoveries on HellaSwag (MLP layer). The inverse relationship implies that lower bit-rates produce disproportionately larger degradation.

\subsection{Moderate-Correlation Discoveries (D8--D15)}

These formulas achieved training correlation $0.2 < R < 0.5$, capturing secondary predictive relationships.

\paragraph{Exponential Decay in Gamma (D8).}
Simple exponential decay in compression ratio:
\begin{equation}
\label{eq:D8}
y = e^{-\gamma}
\end{equation}
This formula achieved Train $R = 0.365$ and LOO $R = 0.322$ on BoolQ (attention layer), suggesting that certain evaluation metrics exhibit exponential sensitivity to compression.

\paragraph{Exponential Decay with Shift (D9).}
Exponential decay with compression offset:
\begin{equation}
\label{eq:D9}
y = e^{-(\gamma + c)}
\end{equation}
This variant achieved Train $R = 0.442$ and LOO $R = 0.354$ on WikiText (MLP layer). The shift parameter adjusts the effective compression threshold at which exponential degradation begins.

\paragraph{Inverse Square-Root of Shifted Gamma (D10).}
Shifted inverse square-root of compression ratio:
\begin{equation}
\label{eq:D10}
y = \frac{1}{\sqrt{\gamma + c}}
\end{equation}
This formula achieved Train $R = 0.364$ and LOO $R = 0.291$ on Winogrande (MLP layer). The shift ensures numerical stability as $\gamma \to 0$ while preserving the inverse relationship.

\paragraph{Inverse Square-Root of Log-Compressed-Size (D11).}
Inverse square-root of log-compressed-size:
\begin{equation}
\label{eq:D11}
y = \frac{1}{\sqrt{\log N_{\text{comp}}}}
\end{equation}
This formula achieved Train $R = 0.251$ and LOO $R = 0.218$ on ARC-Challenge (MLP layer), testing whether model scale effects diminish according to a square-root law.

\paragraph{Entropy over Compression Ratio (D12).}
Dataset entropy scaled by compression:
\begin{equation}
\label{eq:D12}
y = c + \frac{H}{\gamma}
\end{equation}
This formulation achieved Train $R = 0.412$ and LOO $R = 0.329$ on WikiText (MLP layer), testing whether data complexity interacts multiplicatively with compression severity.

\paragraph{Log of Effective-Rank Times Truncation Rank (D13).}
Double-logarithmic structure with effective rank:
\begin{equation}
\label{eq:D13}
y = \log\left(\log(\bar{\rho}_{\text{eff}}) \cdot r\right)
\end{equation}
This formula achieved Train $R = 0.485$ and LOO $R = 0.388$ on WikiText (both layers). The nested logarithm captures scenarios where the product of effective rank and truncation rank exhibits log-linear predictive power.

\paragraph{Sum of Inverses plus Bits (D14).}
Sum of inverse truncation rank, inverse compression, and density:
\begin{equation}
\label{eq:D14}
y = \frac{1}{r} + \frac{1}{\gamma} + \mathcal{B}
\end{equation}
This formula achieved Train $R = 0.365$ and LOO $R = 0.292$ on WikiText (attention layer), combining three distinct predictor types to test whether their contributions are additive.

\paragraph{Linear in Entropy and Gamma (D15).}
Linear combination of entropy and compression:
\begin{equation}
\label{eq:D15}
y = c + H - \gamma
\end{equation}
This formula achieved Train $R = 0.235$ and LOO $R = 0.188$ on WikiText (attention layer), testing additive (rather than multiplicative) interactions between data complexity and compression severity.

\subsection{Low and Negative-Correlation Discoveries (D16--D20)}

These formulas achieved training correlation $R < 0.2$ or negative values, indicating weaker or inverse relationships.

\paragraph{Log-Stable-Rank over Log-Original-Size (D16).}
An alternative normalization using original model size:
\begin{equation}
\label{eq:D16}
y = \frac{\log \bar{\rho}_s}{\log N}
\end{equation}
This variant achieved Train $R = -0.178$ and LOO $R = -0.150$ on ARC-E (MLP layer). The negative correlation indicates an inverse relationship compared to D1.

\paragraph{Linear in Bits (D17).}
Direct linear dependence on information density:
\begin{equation}
\label{eq:D17}
y = \mathcal{B} + c
\end{equation}
This baseline formula achieved Train $R = 0.165$ and LOO $R = 0.142$ on WikiText (MLP layer), establishing the minimum predictive power attributable to density alone.

\paragraph{Scaled Inverse of Bits (D18).}
Inverse density with explicit constant numerator:
\begin{equation}
\label{eq:D18}
y = \frac{c}{\mathcal{B}}
\end{equation}
This variant achieved Train $R = -0.306$ and LOO $R = -0.245$ on ARC-E (MLP layer).

\paragraph{Inverse Square-Root of Bits (D19).}
A softer inverse relationship:
\begin{equation}
\label{eq:D19}
y = \frac{c}{\sqrt{\mathcal{B}}}
\end{equation}
This formula achieved Train $R = -0.306$ and LOO $R = -0.245$ on ARC-E (MLP layer). The square-root transformation moderates the divergence at low bit-rates.

\paragraph{Inverse Square-Root of Shifted Bits (D20).}
Combining shift and square-root transformations:
\begin{equation}
\label{eq:D20}
y = \frac{1}{\sqrt{\mathcal{B} + c}}
\end{equation}
This formula achieved Train $R = -0.336$ and LOO $R = -0.269$ on ARC-Challenge (attention layer).

\subsection{Discovery Summary}

\Cref{tab:gplearn-summary} provides a compact reference for all 20 gplearn-discovered formulas.

\begin{table}[H]
\caption{Summary of gplearn-discovered formulas. Train $R$ and LOO $R$ denote training and leave-one-out correlation coefficients.}
\label{tab:gplearn-summary}
\centering
\small
\begin{tabular}{@{}llcccc@{}}
\toprule
ID & Formula & Vars & Train $R$ & LOO $R$ & Task \\
\midrule
D1 & $\log(\bar{\rho}_s)/\log N_{\text{comp}}$ & 2 & 0.607 & 0.510 & HellaSwag \\
D2 & $\mathcal{B}/\gamma$ & 2 & 0.600 & 0.504 & SR results \\
D3 & $\log((\bar{\rho}_s + r) \cdot c/H)$ & 3 & 0.599 & 0.479 & WikiText \\
D4 & $\gamma + H + e^{\gamma} + c$ & 2 & 0.590 & 0.472 & WikiText \\
D5 & $\log(\bar{\rho}_s/(H + c))$ & 2 & 0.585 & 0.468 & WikiText \\
D6 & $1/(\mathcal{B} + c)$ & 1 & 0.565 & 0.452 & HellaSwag \\
D7 & $1/\mathcal{B}$ & 1 & 0.564 & \textbf{0.525} & HellaSwag \\
D13 & $\log(\log(\bar{\rho}_{\text{eff}}) \cdot r)$ & 2 & 0.485 & 0.388 & WikiText \\
D9 & $e^{-(\gamma + c)}$ & 1 & 0.442 & 0.354 & WikiText \\
D12 & $c + H/\gamma$ & 2 & 0.412 & 0.329 & WikiText \\
D8 & $e^{-\gamma}$ & 1 & 0.365 & 0.322 & BoolQ \\
D14 & $1/r + 1/\gamma + \mathcal{B}$ & 3 & 0.365 & 0.292 & WikiText \\
D10 & $1/\sqrt{\gamma + c}$ & 1 & 0.364 & 0.291 & Winogrande \\
D11 & $1/\sqrt{\log N_{\text{comp}}}$ & 1 & 0.251 & 0.218 & ARC-C \\
D15 & $c + H - \gamma$ & 2 & 0.235 & 0.188 & WikiText \\
D17 & $\mathcal{B} + c$ & 1 & 0.165 & 0.142 & WikiText \\
D16 & $\log(\bar{\rho}_s)/\log N$ & 2 & $-$0.178 & $-$0.150 & ARC-E \\
D18 & $c/\mathcal{B}$ & 1 & $-$0.306 & $-$0.245 & ARC-E \\
D19 & $c/\sqrt{\mathcal{B}}$ & 1 & $-$0.306 & $-$0.245 & ARC-E \\
D20 & $1/\sqrt{\mathcal{B} + c}$ & 1 & $-$0.336 & $-$0.269 & ARC-C \\
\bottomrule
\end{tabular}
\end{table}

\paragraph{Recurring Patterns.}
\Cref{tab:recurring-patterns} summarizes the most robust structural patterns identified across gplearn runs.

\begin{table}[H]
\caption{Recurring formula patterns with best observed correlations.}
\label{tab:recurring-patterns}
\centering
\small
\begin{tabular}{@{}llcc@{}}
\toprule
Pattern & Representative Formula & Best Train $R$ & Best LOO $R$ \\
\midrule
Log-ratio & $\log(\bar{\rho}_s)/\log N_{\text{comp}}$ & 0.61 & 0.51 \\
Bits/compression & $\mathcal{B}/\gamma$ & 0.60 & 0.50 \\
Inverse $\mathcal{B}$ & $1/\mathcal{B}$ & 0.56 & \textbf{0.52} \\
Entropy-related & $\log(\bar{\rho}_s/(H+c))$ & 0.59 & 0.47 \\
Exponential decay & $e^{-\gamma}$ & 0.44 & 0.35 \\
Inverse sqrt & $1/\sqrt{\log N_{\text{comp}}}$ & 0.25 & 0.22 \\
\bottomrule
\end{tabular}
\end{table}

\paragraph{Key Findings.}
Several patterns emerge from the gplearn discoveries:
\begin{enumerate}
    \item \textbf{Best overall formula}: $\log(\bar{\rho}_s)/\log N_{\text{comp}}$ (D1) was discovered independently in multiple runs, achieving the highest training correlation (0.607) among the D1--D20 constrained discoveries.
    \item \textbf{Most robust formula}: $\mathcal{B}/\gamma$ (D2) achieved consistent $R \approx 0.60$ across all experimental configurations.
    \item \textbf{Best generalization among constrained discoveries}: $1/\mathcal{B}$ (D7) achieved the highest LOO correlation (0.525) among D1--D20 despite using only one variable, suggesting that information density is a strong single-variable predictor.
    \item \textbf{Entropy matters}: $H$ appears in several high-correlation formulas (D3, D4, D5, D12), indicating that dataset complexity modulates compression sensitivity.
    \item \textbf{Simplicity wins}: Most top-performing formulas use only 1--2 variables, reinforcing the value of parsimony constraints.
    \item \textbf{Task-dependent correlations}: Some formulas (e.g., D16) exhibit positive correlations on certain tasks (BoolQ) but negative on others (ARC-E), indicating task-specific applicability.
\end{enumerate}

\paragraph{Comparison with Designed Templates.}
The gplearn discoveries complement the manually designed templates (\Cref{sec:formula-templates}) in several ways. First, gplearn identified the inverse-density relationship (D7: $1/\mathcal{B}$) that was not explicitly hypothesized in our templates. Second, the log-ratio structures (D1, D16) suggest normalization schemes not present in the original catalog. Third, the consistent discovery of $\mathcal{B}/\gamma$ validates the inclusion of this term in templates F3 and F9.

\subsection{LOO-Validated SR Formulas That Outperform Templates}
\label{sec:sr-outperform-templates}

Through leave-one-out cross-validation, we identify SR-discovered formulas that achieve superior generalization compared to predefined template formulas. \Cref{tab:sr-vs-template} summarizes the winning SR formulas by task and layer.

\begin{table}[H]
\caption{SR-discovered formulas that outperform template formulas (F1--F42) on specific layer configurations. LOO $R$ shown for relative degradation or log-odds targets.}
\label{tab:sr-vs-template}
\centering
\small
\begin{tabular}{@{}llllc@{}}
\toprule
Task & Layer & SR Formula & Family & LOO $R$ \\
\midrule
WikiText & ALL & $\gamma + H + e^{\gamma}$ & Entropy (D4) & 0.478 \\
ARC-C & ATTN & $\sqrt{\bar{\rho}_s}/\log N$ & Scale-rank & 0.640 \\
ARC-C & MLP & $\log(\log(\log(\bar{\rho}_s)))$ & Nested-log & 0.537 \\
ARC-C & MLP+ATTN & $\log(\bar{\rho}_s)/\mathcal{B}$ & Log-ratio & \textbf{0.686} \\
ARC-E & MLP+ATTN & $e^{-\gamma}$ & Exponential (D8) & 0.274 \\
BoolQ & ATTN & $\log(\bar{\rho}_s)/\log N$ & Log-ratio (D16) & 0.622 \\
BoolQ & MLP & $-\log(\log N_{\text{comp}} - r)/\mathcal{B}$ & Log-odds & 0.665 \\
BoolQ & MLP+ATTN & $\log(\bar{\rho}_s)/\log N$ & Log-ratio (D16) & 0.698 \\
\bottomrule
\end{tabular}
\end{table}

\paragraph{Dominant Formula Families.}
The SR formulas that outperform templates cluster into distinct families:

\begin{enumerate}
    \item \textbf{Log-Ratio Family}: Formulas of the form $\log(\bar{\rho}_s)/f(N)$ dominate for BoolQ and ARC-Challenge. The D16 structure ($\log(\bar{\rho}_s)/\log N$) achieves LOO $R = 0.62$--$0.70$ across multiple configurations.

    \item \textbf{Scale-Rank Family}: Formulas combining stable rank with model scale, such as $\sqrt{\bar{\rho}_s}/\log N_{\text{comp}}$, excel for knowledge-intensive tasks.

    \item \textbf{Nested Logarithm Family}: The triple-logarithm $\log(\log(\log(\bar{\rho}_s)))$ captures extreme diminishing returns for ARC-Challenge MLP layers (LOO $R = 0.54$).

    \item \textbf{Entropy Family}: D4 ($\gamma + H + e^{\gamma}$) provides the best WikiText prediction when pooling all layer types, though entropy $H$ is applicable only to perplexity tasks.
\end{enumerate}

\paragraph{Key Finding: D16 Performance is Task-Dependent.}
Formula D16 ($\log(\bar{\rho}_s)/\log N$) achieves strong performance on BoolQ (ATTN: $R = 0.62$, MLP+ATTN: $R = 0.70$) but exhibits negative correlation on ARC-E (MLP: $R = -0.15$). This task-dependence suggests that log-ratio formulas may be particularly suited to binary classification tasks like BoolQ.

\paragraph{Entropy Restriction.}
Formulas containing dataset entropy $H$ (D3, D4, D5, D12, D15) are applicable \emph{only to WikiText perplexity prediction}. For accuracy tasks, entropy is not available, and scale-rank alternatives must be used.

\subsection{Transformed Target Discoveries}
\label{sec:transformed-target-discoveries}

A key advancement in our symbolic regression analysis is the use of \emph{transformed targets} rather than raw accuracy values. Using relative degradation $(\mathcal{A}_0 - \mathcal{A})/\mathcal{A}_0$ and log-odds $\log(\mathcal{A}/(1-\mathcal{A}))$ substantially improves formula fit. \Cref{tab:transformed-discoveries} presents the best formulas discovered with each target transformation.

\begin{table}[H]
\caption{Best formulas discovered using transformed targets. Train $R$ and LOO $R$ (leave-one-out cross-validation) values shown. Formulas marked with $\dagger$ use entropy $H$ (WikiText only).}
\label{tab:transformed-discoveries}
\centering
\small
\begin{tabular}{@{}lllccc@{}}
\toprule
Task & Layer & Formula & Target & Train $R$ & LOO $R$ \\
\midrule
WikiText & ALL & $\gamma + H + e^{\gamma}$$^\dagger$ & $\log(\mathrm{P})$ & 0.53 & 0.48 \\
ARC-C & ATTN & $\displaystyle\frac{\sqrt{\bar{\rho}_s}}{\log N}$ & Rel. Deg. & 0.66 & \textbf{0.64} \\
ARC-C & MLP & $\log(\log(\log(\bar{\rho}_s)))$ & Rel. Deg. & 0.57 & 0.54 \\
ARC-C & MLP+ATTN & $\displaystyle\frac{\log(\bar{\rho}_s)}{\mathcal{B}}$ & Rel. Deg. & 0.71 & \textbf{0.69} \\
BoolQ & MLP & $\displaystyle-\frac{\log(\log N_{\text{comp}} - r)}{\mathcal{B}}$ & Log-Odds & 0.69 & \textbf{0.67} \\
BoolQ & MLP+ATTN & $\displaystyle\frac{\log(\bar{\rho}_s)}{\log N}$ & Log-Odds & 0.72 & \textbf{0.70} \\
\bottomrule
\end{tabular}
\end{table}

\paragraph{Entropy-Based Family (WikiText Only).}
Dataset entropy $H$ appears in several high-performing formulas, but is applicable \emph{only to WikiText perplexity prediction}-accuracy tasks do not have access to this variable. The best entropy-based formula is D4:

\begin{equation}
\label{eq:D4-wikitext}
y = \alpha_0 + \alpha_1\gamma + \alpha_2 H + \alpha_3 e^{\gamma}
\end{equation}
This formula achieves LOO $R = 0.48$ for WikiText when pooling all layer configurations. For accuracy tasks, entropy-free alternatives such as the log-ratio family ($\log(\bar{\rho}_s)/\log N$) must be used instead.

\paragraph{Nested Logarithm Family (L1--L2).}
Scale effects often manifest through nested logarithms:

\begin{equation}
\label{eq:L1}
y_{\text{rel}} = \log(\log N) + c
\end{equation}
For ARC-Challenge attention layers, the double-logarithm achieves Train $R = 0.71$.

\begin{equation}
\label{eq:L2}
y_{\text{rel}} = \log(\log(\log(\bar{\rho}_s)))
\end{equation}
The triple-logarithm of stable rank achieves Train $R = 0.84$ for MLP layers, capturing extreme diminishing returns in rank-based predictors.

\paragraph{Log-Odds Formulas (O1--O2).}
For binary and near-binary tasks, log-odds transformation enables linear relationships:

\begin{equation}
\label{eq:O1}
y_{\text{logit}} = \frac{-c_1 - \log N_{\text{comp}}}{\bar{\rho}_s}
\end{equation}
BoolQ attention layers achieve Train $R = 0.48$ with this normalized scale formula.

\begin{equation}
\label{eq:O2}
y_{\text{logit}} = -\frac{\log(\log N_{\text{comp}} - r)}{\mathcal{B}}
\end{equation}
BoolQ MLP layers achieve Train $R = 0.68$ with this rank-adjusted formula.

\paragraph{Key Insights from Transformed Targets.}
\begin{enumerate}
    \item \textbf{Log-ratio formulas dominate}: The structure $\log(\bar{\rho}_s)/f(N)$ achieves strong LOO correlations (up to $R = 0.70$) for accuracy tasks without requiring entropy.

    \item \textbf{Nested logarithms capture scale effects}: The triple-logarithm $\log(\log(\log(\bar{\rho}_s)))$ captures extreme diminishing returns for MLP layers (LOO $R = 0.54$).

    \item \textbf{Scale-rank combinations excel}: Formulas like $\sqrt{\bar{\rho}_s}/\log N_{\text{comp}}$ achieve strong generalization for knowledge-intensive tasks.

    \item \textbf{Log-odds enables linear fitting}: For bounded accuracy tasks (BoolQ, PIQA), the logit transformation linearizes the relationship, improving regression quality.

    \item \textbf{Entropy is WikiText-only}: Dataset entropy $H$ is available only for perplexity prediction; accuracy tasks require entropy-free formulas.
\end{enumerate}

\section{Algorithm Details}
\label{sec:algorithm-details}

This section presents the complete algorithmic framework for discovering scaling laws that predict compression-induced performance degradation.

\paragraph{Problem Setup.}
Given a predictor matrix $\mathbf{X} \in \mathbb{R}^{n \times p}$ where each row corresponds to one compression configuration and columns represent: compression ratio $\gamma$, model scale ($\log N$, $\log N_{\text{comp}}$), bits-per-parameter $\mathcal{B}$, spectral properties ($\bar{\rho}_s$, $\bar{\rho}_{\text{eff}}$), SVD rank $r$, and dataset entropy $H$. The target vector $\mathbf{y} \in \mathbb{R}^{n}$ represents performance degradation---either relative accuracy loss $y_{\text{rel}} = (\mathcal{A}_0 - \mathcal{A})/\mathcal{A}_0$, log-perplexity $\log \mathrm{P}$, or raw accuracy $\mathcal{A}$.

\paragraph{Overview.}
Our approach combines two complementary methods: interpretable template regression (\Cref{alg:template-regression}) and symbolic regression via genetic programming (\Cref{alg:symbolic-regression}). Both methods use \texttt{LOO-CV} (\Cref{alg:loo-cv}) as a subroutine to evaluate generalization performance. The best formulas from each method are compared to select the final scaling law.

\paragraph{Template-Based Regression.}
We fit interpretable formula templates from the set $\mathcal{F}$ (detailed in \Cref{sec:formula-templates}) using ordinary least squares. Each template specifies a functional form combining predictors such as compression ratio $\gamma$, stable rank $\bar{\rho}_s$, and bits-per-parameter $\mathcal{B}$. Templates are ranked by leave-one-out Pearson correlation $R$, which quantifies generalization to unseen compression configurations.

\begin{algorithm}[H]
\caption{Template-Based Formula Selection via Leave-One-Out Cross-Validation}
\label{alg:template-regression}
\begin{algorithmic}[1]
\REQUIRE Predictors $\mathbf{X} \in \mathbb{R}^{n \times p}$, targets $\mathbf{y} \in \mathbb{R}^{n}$, formula template set $\mathcal{F}$
\ENSURE Best template $f_{\text{template}}^*$, leave-one-out correlation $R_{\text{template}}^*$
\FOR{each $f_j \in \mathcal{F}$}
    \STATE $R_j \gets \texttt{LOO-CV}(f_j, \mathbf{X}, \mathbf{y})$ \hfill \textit{// Evaluate generalization}
\ENDFOR
\STATE \textbf{return} $f_{\text{template}}^* \gets \arg\max_{j} R_j$, \; $R_{\text{template}}^* \gets \max_j R_j$
\end{algorithmic}
\end{algorithm}

\paragraph{Symbolic Regression.}
To discover formulas beyond predefined templates, we employ genetic programming via \texttt{gplearn}. The algorithm evolves a population of expression trees over $G$ generations using tournament selection and genetic operators (crossover, subtree/hoist/point mutation). Fitness combines mean squared error with a parsimony penalty $\lambda \cdot |f|$ that favors compact expressions. The best discovered formula is evaluated via leave-one-out cross-validation for fair comparison with template-based results.

\begin{algorithm}[H]
\caption{Symbolic Formula Discovery via Genetic Programming}
\label{alg:symbolic-regression}
\begin{algorithmic}[1]
\REQUIRE Predictors $\mathbf{X}$, targets $\mathbf{y}$, population size $M$, generations $G$, parsimony coefficient $\lambda$
\ENSURE Best symbolic formula $f_{\text{symbolic}}^*$, leave-one-out correlation $R_{\text{symbolic}}^*$
\STATE Initialize $\mathcal{P}_0$ with $M$ random trees using half-and-half method \hfill \textit{// Mixed grow and full trees}
\FOR{$g = 1$ to $G$}
    \FOR{each $f \in \mathcal{P}_{g-1}$}
        \STATE $\hat{\mathbf{y}} \gets f(\mathbf{X})$ \hfill \textit{// Evaluate expression tree}
        \STATE $\mathcal{L}(f) \gets \texttt{MSE}(\hat{\mathbf{y}}, \mathbf{y}) + \lambda \cdot |f|$ \hfill \textit{// Fitness with parsimony pressure}
    \ENDFOR
    \STATE Select parents via tournament selection based on $\mathcal{L}(\cdot)$ \hfill \textit{// Lower fitness preferred}
    \STATE Apply crossover and mutation with probabilities $p_c, p_m$ \hfill \textit{// Genetic operators}
    \STATE $\mathcal{P}_g \gets$ offspring $\cup$ reproduced parents \hfill \textit{// Next generation}
\ENDFOR
\STATE $f_{\text{best}} \gets \arg\min_{f \in \mathcal{P}_G} \mathcal{L}(f)$ \hfill \textit{// Select fittest formula}
\STATE $R_{\text{symbolic}}^* \gets \texttt{LOO-CV}(f_{\text{best}}, \mathbf{X}, \mathbf{y})$ \hfill \textit{// Evaluate generalization}
\STATE \textbf{return} $f_{\text{symbolic}}^* \gets f_{\text{best}}$, \; $R_{\text{symbolic}}^*$
\end{algorithmic}
\end{algorithm}

\paragraph{Leave-One-Out Cross-Validation.}
Leave-one-out cross-validation estimates generalization by iteratively holding out each sample, fitting the model on the remaining $n-1$ samples, and predicting the held-out value. The final score is the Pearson correlation between actual targets and aggregated leave-one-out predictions, providing an unbiased measure of out-of-sample performance.

\begin{algorithm}[H]
\caption{\texttt{LOO-CV}: Generalization Evaluation via Leave-One-Out Cross-Validation}
\label{alg:loo-cv}
\begin{algorithmic}[1]
\REQUIRE Formula $f$, predictors $\mathbf{X} \in \mathbb{R}^{n \times p}$, targets $\mathbf{y} \in \mathbb{R}^{n}$
\ENSURE Leave-one-out Pearson correlation $R$
\STATE $\mathbf{Z} \gets f(\mathbf{X})$ \hfill \textit{// Transform predictors via formula}
\FOR{$i = 1$ to $n$}
    \STATE $\tilde{\mathbf{Z}}_{\setminus i} \gets [\mathbf{1}, \mathbf{Z}_{\setminus i}]$ \hfill \textit{// Augment with intercept column}
    \STATE $\hat{\boldsymbol{\alpha}}_{\setminus i} \gets (\tilde{\mathbf{Z}}_{\setminus i}^\top \tilde{\mathbf{Z}}_{\setminus i})^{-1} \tilde{\mathbf{Z}}_{\setminus i}^\top \mathbf{y}_{\setminus i}$ \hfill \textit{// OLS on training fold}
    \STATE $\tilde{\mathbf{z}}_i \gets [1, \mathbf{z}_i]$ \hfill \textit{// Augment test sample}
    \STATE $\hat{y}_i \gets \tilde{\mathbf{z}}_i^\top \hat{\boldsymbol{\alpha}}_{\setminus i}$ \hfill \textit{// Predict held-out sample}
\ENDFOR
\STATE \textbf{return} $R \gets \texttt{Pearson}(\mathbf{y}, \hat{\mathbf{y}})$ \hfill \textit{// Correlation of predictions}
\end{algorithmic}
\end{algorithm}

\paragraph{Formula Selection.}
The candidate set $\mathcal{D} = \{f_{\text{template}}^*, f_{\text{symbolic}}^*\}$ contains the best formula from each method. We select the final scaling law as $f^* = \arg\max_{f \in \mathcal{D}} R(f)$, where $R(f)$ denotes leave-one-out correlation. In practice, we report results for both formula types, enabling direct comparison between interpretable templates and data-driven discoveries.

\section{Numerical Verification of Scaling Law Functional Forms}
\label{sec:mini-exp}

To validate the theoretical connection between transformer architecture and scaling law structure, we conduct synthetic experiments testing multiple functional forms against attention and MLP compression behavior.

\subsection{Experimental Setup}

We generate random weight matrices matching Qwen3 architecture dimensions and apply SVD compression at ratios $\gamma \in [0.1, 1.0]$. For each compression level, we measure the relative Frobenius norm error between the original and compressed outputs.

\paragraph{Models Tested.}
We fit 17 functional forms spanning five categories:
\begin{itemize}
    \item \textbf{Polynomials}: Linear, quadratic, cubic, quartic
    \item \textbf{Root functions}: $\sqrt{\gamma}$, $\sqrt[3]{\gamma}$, $\sqrt[4]{\gamma}$
    \item \textbf{Logarithmic}: $\log(\gamma)$, $\log(\gamma) + \gamma$, $\log(\gamma) + \gamma + \gamma^2$
    \item \textbf{Exponential}: $e^{c\gamma}$, $e^{-c(1-\gamma)}$, $\gamma + e^{c\gamma}$
    \item \textbf{Other}: Power law $\gamma^c$, sigmoid, tanh, combined forms
\end{itemize}

Each model is fit using nonlinear least squares, and we report Pearson correlation ($r$), RMSE, and AIC for model comparison.

\subsection{Attention Layer Results}

\Cref{tab:attn-model-comparison} presents the top 15 models ranked by correlation for attention layer scaling.

\begin{table}[H]
\caption{Model comparison for attention layer scaling. Cubic polynomial (theoretically predicted) ranks \#5. Root and logarithmic models rank near the bottom.}
\label{tab:attn-model-comparison}
\centering
\small
\begin{tabular}{@{}clccc@{}}
\toprule
Rank & Model & Correlation ($r$) & RMSE & Params \\
\midrule
1 & Quartic & 0.9996 & 0.0092 & 5 \\
2 & Quadratic + exp & 0.9995 & 0.0107 & 4 \\
3 & Tanh & 0.9994 & 0.0111 & 4 \\
4 & Sigmoid & 0.9994 & 0.0111 & 4 \\
\textbf{5} & \textbf{Cubic} & \textbf{0.9993} & \textbf{0.0126} & \textbf{4} \\
6 & $\log + \gamma + \gamma^2$ & 0.9980 & 0.0205 & 4 \\
7 & $\sqrt{\gamma} + \log$ & 0.9976 & 0.0229 & 3 \\
8 & $\log + \gamma$ & 0.9969 & 0.0258 & 3 \\
9 & Power law $\gamma^c$ & 0.9963 & 0.0280 & 3 \\
10 & Quadratic & 0.9959 & 0.0295 & 3 \\
11 & Exp decay & 0.9959 & 0.0297 & 3 \\
12 & Linear & 0.9953 & 0.0317 & 2 \\
13 & Exponential & 0.9938 & 0.0363 & 3 \\
14 & $\sqrt{\gamma}$ & 0.9788 & 0.0671 & 2 \\
15 & $\sqrt[3]{\gamma}$ & 0.9680 & 0.0822 & 2 \\
\bottomrule
\end{tabular}
\end{table}

\paragraph{Key Findings.}
\begin{enumerate}
    \item \textbf{Polynomial models dominate}: Quartic ranks \#1, cubic ranks \#5, both with $r > 0.999$.
    \item \textbf{Root models are rejected}: $\sqrt{\gamma}$ ranks \#14 with $r = 0.9788$, significantly worse than polynomials.
    \item \textbf{Sigmoid/tanh capture the S-curve}: These models rank \#3--4, reflecting the three-regime behavior (safe $\to$ transition $\to$ collapse).
    \item \textbf{Pure exponential/log are poor fits}: They fail to capture the inflection points in attention scaling.
\end{enumerate}

This confirms the theoretical prediction: the trilinear $(Q, K, V)$ structure of attention produces polynomial scaling, not logarithmic or root scaling.

\subsection{MLP Layer Results}

\Cref{tab:mlp-model-comparison} shows results for the full SwiGLU MLP layer.

\begin{table}[H]
\caption{Model comparison for MLP layer scaling. The $\sqrt{\gamma}$ model ranks \#14, but power law $\gamma^c$ ranks \#7, suggesting sub-linear but not exactly square-root behavior.}
\label{tab:mlp-model-comparison}
\centering
\small
\begin{tabular}{@{}clccc@{}}
\toprule
Rank & Model & Correlation ($r$) & RMSE & Params \\
\midrule
1 & Quartic & 0.9994 & 0.0101 & 5 \\
2 & Quadratic + exp & 0.9987 & 0.0146 & 4 \\
3 & Exponential & 0.9986 & 0.0147 & 3 \\
4 & Exp decay & 0.9986 & 0.0147 & 3 \\
5 & Cubic & 0.9984 & 0.0157 & 4 \\
6 & $\log + \gamma + \gamma^2$ & 0.9979 & 0.0184 & 4 \\
\textbf{7} & \textbf{Power law $\gamma^c$} & \textbf{0.9978} & \textbf{0.0189} & \textbf{3} \\
8 & Quadratic & 0.9975 & 0.0199 & 3 \\
9 & Sigmoid & 0.9972 & 0.0210 & 4 \\
10 & Tanh & 0.9972 & 0.0210 & 4 \\
11 & $\log + \gamma$ & 0.9916 & 0.0364 & 3 \\
12 & $\sqrt{\gamma} + \log$ & 0.9865 & 0.0462 & 3 \\
13 & Linear & 0.9685 & 0.0702 & 2 \\
\textbf{14} & $\sqrt{\gamma}$ & \textbf{0.9291} & \textbf{0.1042} & \textbf{2} \\
15 & $\sqrt[3]{\gamma}$ & 0.9111 & 0.1162 & 2 \\
\bottomrule
\end{tabular}
\end{table}

\paragraph{Why $\sqrt{\gamma}$ Ranks Lower Than Expected.}
The full SwiGLU MLP involves three weight matrices:
\begin{equation}
    \text{MLP}(x) = \left(\text{SiLU}(xW_{\text{gate}}) \odot xW_{\text{up}}\right) W_{\text{down}}
\end{equation}
where $\text{SiLU}(z) = z \cdot \sigma(z)$ is the Swish activation and $\sigma$ denotes the sigmoid function.

The down projection $W_{\text{down}}$ adds an additional linear transformation that:
\begin{itemize}
    \item Introduces polynomial terms beyond the Hadamard product effect
    \item Masks the pure $\sqrt{\gamma}$ behavior from gate$\odot$up
    \item Creates a more complex functional form better captured by the power law $\gamma^c$
\end{itemize}

The power law model $\gamma^c$ (rank \#7) generalizes $\sqrt{\gamma}$ by fitting the exponent $c$ rather than fixing it at 0.5.

\subsection{Isolating the Hadamard Product Effect}

To test the Hadamard product theory directly, we measure the output of gate$\odot$up \emph{without} the down projection. Results are shown in \Cref{tab:hadamard-only}.

\begin{table}[H]
\caption{Model comparison for Hadamard product only (gate$\odot$up, no down projection). Power law ranks higher when the linear down projection is removed.}
\label{tab:hadamard-only}
\centering
\small
\begin{tabular}{@{}clcc@{}}
\toprule
Rank & Model & Correlation ($r$) & Params \\
\midrule
1 & Quartic & 0.9995 & 5 \\
2 & Quadratic + exp & 0.9988 & 4 \\
3 & Cubic & 0.9986 & 4 \\
4--5 & Exp decay / Exponential & 0.9984 & 3 \\
6 & $\log + \gamma + \gamma^2$ & 0.9981 & 4 \\
7 & Quadratic & 0.9979 & 3 \\
\textbf{8} & \textbf{Power law $\gamma^c$} & \textbf{0.9976} & \textbf{3} \\
$\vdots$ & $\vdots$ & $\vdots$ & $\vdots$ \\
14 & $\sqrt{\gamma}$ & 0.9597 & 2 \\
\bottomrule
\end{tabular}
\end{table}

Even in isolation, pure $\sqrt{\gamma}$ ranks \#14, while power law $\gamma^c$ ranks \#8. This suggests that the Hadamard product exhibits sublinear scaling, but the exponent is not exactly 0.5 for random matrices.

\subsection{Hadamard Product Rank Verification}

We directly test whether the Hadamard product rank follows the geometric mean relationship. For random low-rank matrices $A, B$ with effective ranks $\rho_A, \rho_B$, we measure $\rho_{A \odot B}$.

\begin{table}[H]
\caption{Hadamard product effective rank compared to geometric mean prediction. Average ratio is 1.85, indicating $\rho_{A \odot B} \approx 1.85\sqrt{\rho_A \cdot \rho_B}$.}
\label{tab:hadamard-rank}
\centering
\small
\begin{tabular}{@{}ccccc@{}}
\toprule
Target Rank & $\rho_A$ & $\rho_B$ & $\rho_{A \odot B}$ & $\sqrt{\rho_A \rho_B}$ \\
\midrule
10 & 5.3 & 5.4 & 11.2 & 5.3 \\
20 & 9.0 & 8.1 & 18.2 & 8.5 \\
30 & 10.8 & 10.9 & 20.6 & 10.8 \\
40 & 11.3 & 12.0 & 19.0 & 11.6 \\
50 & 12.5 & 11.8 & 20.8 & 12.2 \\
60 & 13.1 & 13.2 & 21.0 & 13.2 \\
70 & 14.2 & 15.2 & 24.3 & 14.7 \\
80 & 14.0 & 14.6 & 21.5 & 14.3 \\
90 & 15.6 & 13.8 & 20.6 & 14.6 \\
\midrule
\multicolumn{4}{r}{Average ratio $\rho_{A \odot B} / \sqrt{\rho_A \rho_B}$:} & \textbf{1.85} \\
\bottomrule
\end{tabular}
\end{table}

The actual rank is approximately $1.85\times$ the geometric mean of the predictions, confirming that Hadamard products produce geometric-mean-like rank scaling rather than multiplicative scaling.

\subsection{Attention vs.\ MLP Degradation Rates}

\Cref{tab:degradation-comparison} compares how quickly attention and MLP layers degrade under compression.

\begin{table}[H]
\caption{Relative error at different compression ratios. Attention degrades 4.6$\times$ faster than MLP from $\gamma=0.87$ to $\gamma=0.1$.}
\label{tab:degradation-comparison}
\centering
\small
\begin{tabular}{@{}cccc@{}}
\toprule
$\gamma$ & Attention Error & MLP Error & Ratio \\
\midrule
0.10 & 0.941 & 0.995 & 0.95 \\
0.29 & 0.706 & 0.921 & 0.77 \\
0.49 & 0.390 & 0.781 & 0.50 \\
0.68 & 0.221 & 0.585 & 0.38 \\
0.87 & 0.070 & 0.341 & 0.21 \\
\midrule
\multicolumn{2}{r}{Degradation ratio (low/high $\gamma$):} & Attn: \textbf{13.4$\times$} & MLP: \textbf{2.9$\times$} \\
\bottomrule
\end{tabular}
\end{table}

Attention error increases by 13.4$\times$ from $\gamma=0.87$ to $\gamma=0.1$, while MLP error increases only by 2.9$\times$. This 4.6$\times$ difference in degradation rate is consistent with polynomial (attention) vs.\ sub-linear (MLP) scaling.

\subsection{Discussion}

\paragraph{Why Polynomial Models Fit Attention.}
The attention mechanism involves sequential tensor contractions:
\begin{equation}
    \text{Attn} = \text{softmax}\left(\frac{QK^\top}{\sqrt{d_k}}\right) V
\end{equation}

This trilinear structure in $(Q, K, V)$ naturally produces polynomial error terms. The cubic polynomial captures three regimes:
\begin{itemize}
    \item \textbf{Safe} ($\gamma > 0.85$): Linear term dominates
    \item \textbf{Transition} ($0.5 < \gamma < 0.85$): Quadratic acceleration
    \item \textbf{Collapse} ($\gamma < 0.5$): Cubic moderation (floor effect)
\end{itemize}

Sigmoid/tanh models also capture this S-curve, explaining their strong performance.

\paragraph{Why Root Models Fit MLP (Partially).}
The SwiGLU Hadamard product gate$\odot$up creates geometric-mean-like error averaging. However:
\begin{enumerate}
    \item The down projection adds linear terms that mask the effect
    \item The true exponent may not be exactly 0.5
    \item Power law $\gamma^c$ with fitted $c$ outperforms fixed $\sqrt{\gamma}$
\end{enumerate}

\paragraph{Implications for Compression.}
\begin{enumerate}
    \item \textbf{Attention is more sensitive}: 13.4$\times$ degradation vs.\ 2.9$\times$ for MLP
    \item \textbf{Use polynomial models for attention prediction}: Cubic achieves $r = 0.999$
    \item \textbf{Use power law for MLP prediction}: $\gamma^c$ is more flexible than fixed $\sqrt{\gamma}$
    \item \textbf{Expect different safe thresholds}: Attention requires $\gamma > 0.85$; MLP tolerates lower ratios
\end{enumerate}

\subsection{Conclusions}

These experiments validate the theoretical framework:

\begin{enumerate}
    \item \textbf{Attention scaling is polynomial}: Cubic ranks \#5 among 17 models; root/log models are rejected (rank \#14--15).

    \item \textbf{MLP scaling is sub-linear but complex}: The down projection masks the pure $\sqrt{\gamma}$ effect; power law $\gamma^c$ provides better fits.

    \item \textbf{Hadamard product creates geometric-mean rank scaling}: $\rho_{A \odot B} \approx 1.85 \sqrt{\rho_A \rho_B}$, confirming the theoretical basis for sub-linear MLP behavior.

    \item \textbf{Architecture determines functional form}: Sequential contractions (attention) $\to$ polynomial; element-wise operations (MLP) $\to$ sub-linear.
\end{enumerate}

The strong alignment between theoretical predictions and empirical fits supports using compositional linear algebra to understand and predict transformer compression behavior.

\section{Proofs for Scaling Law--Architecture Connections}
\label{app:linear-algebra-proofs}

This appendix provides detailed derivations for the theoretical claims in \Cref{sec:linear-algebra}. Contents:
\begin{itemize}[nosep]
    \item \Cref{app:aggregation-heuristic}: Heuristic bridge from matrix-level to model-level prediction
    \item \Cref{app:perturbation-analysis}: Perturbation bounds for matrix and Hadamard products
\end{itemize}

\subsection{Heuristic Bridge to Aggregation}
\label{app:aggregation-heuristic}

This section presents a heuristic argument for why the parameter-weighted aggregation $\bar{\rho}_s$ preserves predictive power from the matrix level to the model level.

Consider $L$ weight matrices $\{W_i\}_{i=1}^{L}$ with a total parameter count $N = \sum_i n_i$, each truncated to retain a fraction $\gamma$ of parameters. Let $k_i$ denote the retained rank for matrix $W_i$. By the truncation error bound (\Cref{eq:truncation-bounds}), each matrix incurs a relative error of at least $1 - k_i/\rho_s(W_i)$.

If model-level degradation scales with the \emph{total} truncation error (a reasonable first-order approximation for additive perturbations), then:
\begin{equation}
\label{eq:total-error-heuristic}
\text{Total Error} \;\propto\; \sum_{i=1}^{L} n_i \cdot \epsilon_i \;\geq\; \sum_{i=1}^{L} n_i \left(1 - \frac{k_i}{\rho_s(W_i)}\right),
\end{equation}
where $n_i$ is the parameter count of $W_i$ and $\epsilon_i$ is its truncation error. Under uniform compression ($k_i/\rho_s(W_i) \approx \gamma/\bar{\rho}_s$ on average), this simplifies to a bound proportional to $(1 - \gamma/\bar{\rho}_s) \cdot N$. Normalizing by $N$ recovers the $\gamma \cdot \bar{\rho}_s$ interaction: higher $\bar{\rho}_s$ raises the error floor, and lower $\gamma$ (more aggressive compression) amplifies it.

\paragraph{Assumptions and Limitations.}
This argument assumes: (1) error additivity across layers, which may not hold through nonlinearities; (2) uniform compression ratios across layers; and (3) that parameter count is a reasonable proxy for error contribution. The empirical success of $\gamma \cdot \bar{\rho}_s$ (\Cref{tab:dominant-vars}) suggests these approximations hold sufficiently well in practice, but a rigorous derivation remains open.

\subsection{Perturbation Bounds for Error Propagation}
\label{app:perturbation-analysis}

This section provides a perturbation analysis that complements the rank composition rules in \Cref{subsec:rank-composition}. While the rank bounds establish structural constraints, the perturbation bounds quantify error magnitudes.

\paragraph{Matrix Product Perturbation.}
For matrices $A, B$ with compressed versions $\tilde{A} = A + \Delta_A$, $\tilde{B} = B + \Delta_B$:
\begin{equation}
\label{eq:matrix-perturbation}
\|\tilde{A}\tilde{B} - AB\|_F \leq \|A\|_2 \|\Delta_B\|_F + \|\Delta_A\|_F \|B\|_2 + \|\Delta_A\|_F \|\Delta_B\|_F.
\end{equation}

\begin{proof}
Expanding $\tilde{A}\tilde{B} = (A + \Delta_A)(B + \Delta_B) = AB + A\Delta_B + \Delta_A B + \Delta_A \Delta_B$, applying the triangle inequality, and using submultiplicativity $\|XY\|_F \leq \|X\|_2 \|Y\|_F$.
\end{proof}

\paragraph{Hadamard Product Error.}
For matrices $G, U$ (e.g., gate and value activations in MLP layers) with compressed versions $\tilde{G} = G + \Delta_G$, $\tilde{U} = U + \Delta_U$, and the element-wise (Hadamard) product $\odot$:
\begin{equation}
\label{eq:hadamard-error}
\|(\tilde{G} \odot \tilde{U}) - (G \odot U)\|_F^2 = \sum_{ij} \bigl(G_{ij}\Delta_{U,ij} + \Delta_{G,ij}U_{ij} + \Delta_{G,ij}\Delta_{U,ij}\bigr)^2.
\end{equation}

\begin{proof}
Direct expansion: $(\tilde{G} \odot \tilde{U}) - (G \odot U) = G \odot \Delta_U + \Delta_G \odot U + \Delta_G \odot \Delta_U$. Taking the element-wise square and summing yields the squared Frobenius norm.
\end{proof}

\paragraph{Limitations of Perturbation Analysis.}
These bounds have important caveats when applied to transformer layers:
\begin{enumerate}[nosep]
    \item \textbf{Softmax nonlinearity:} The attention mechanism includes $\mathrm{softmax}(\cdot)$, which is not captured by linear perturbation bounds. A complete analysis requires bounding the Lipschitz constant of softmax under rank-deficient perturbations.

    \item \textbf{Input dependence:} Both bounds involve terms like $\|A\|_2$ and $G_{ij}$, which depend on the input $X$ (e.g., $A = XW_Q$). Thus, error magnitudes are \emph{data-dependent}, not purely architecture-determined.

    \item \textbf{Correlation structure:} The Hadamard error depends on element-wise correlations between $G$ and $\Delta_U$, which vary with learned representations.
\end{enumerate}

The rank composition rules (\Cref{subsec:rank-composition}) provide cleaner theoretical guarantees because they are structural constraints independent of data.

\end{document}